\documentclass[sigconf]{acmart}
\usepackage{algorithmic}
\usepackage{setspace}

\usepackage{multicol}  
\usepackage{multirow}
\usepackage{booktabs}
\usepackage{amsmath}
\usepackage{caption}
\usepackage[linesnumbered, ruled]{algorithm2e}
\hypersetup{draft}
\usepackage{enumitem}
\usepackage{bbm}
\usepackage{tabularx}
\usepackage{graphicx}
\usepackage{subcaption}

\DeclareMathOperator*{\argmin}{arg\,min}
\newcolumntype{C}{>{\centering\arraybackslash}X}

\AtBeginDocument{%
  \providecommand\BibTeX{{%
    \normalfont B\kern-0.5em{\scshape i\kern-0.25em b}\kern-0.8em\TeX}}}

\copyrightyear{2021}
\acmYear{2021}
\setcopyright{acmcopyright}\acmConference[KDD '21]{Proceedings of the 27th ACM
SIGKDD Conference on Knowledge Discovery and Data Mining}{August 14--18,
2021}{Virtual Event, Singapore}
\acmBooktitle{Proceedings of the 27th ACM SIGKDD Conference on Knowledge
Discovery and Data Mining (KDD '21), August 14--18, 2021, Virtual Event, Singapore}
\acmPrice{15.00}
\acmDOI{10.1145/3447548.3467364}
\acmISBN{978-1-4503-8332-5/21/08}

\begin{document}
\fancyhead{}
\title{NRGNN: Learning a Label Noise-Resistant Graph Neural Network on Sparsely and Noisily Labeled Graphs}

\author{Enyan Dai}
\affiliation{
\institution{The Pennsylvania State University}}
\email{emd5759@psu.edu}
\author{Charu Aggarwal}
\affiliation{IBM}
\email{charu@us.ibm.com}

\author{Suhang Wang}
\affiliation{The Pennsylvania State University}
\email{szw494@psu.edu}



\begin{abstract}
Graph Neural Networks (GNNs) have achieved promising results for semi-supervised learning tasks on graphs such as node classification. Despite the great success of GNNs, many real-world graphs are often sparsely and noisily labeled, which could significantly degrade the performance of GNNs, as the noisy information could propagate to unlabeled nodes via graph structure. Thus, it is important to develop a label noise-resistant GNN for semi-supervised node classification. Though extensive studies have been conducted to learn neural networks with noisy labels, they mostly focus on independent and identically distributed data and assume a large number of noisy labels are available, which are not directly applicable for GNNs.  
Thus, we investigate a novel problem of learning a robust GNN with noisy and limited labels. To alleviate the negative effects of label noise, we propose to link the unlabeled nodes with labeled nodes of high feature similarity to bring more clean label information. Furthermore, accurate pseudo labels could be obtained by this strategy to provide more supervision and further reduce the effects of label noise. Our theoretical and empirical analysis verify the effectiveness of these two strategies under mild conditions. Extensive experiments on real-world datasets demonstrate the effectiveness of the proposed method in learning a robust GNN with noisy and limited labels.
\end{abstract}




\maketitle

\section{Introduction}
Graph structured data is very pervasive in real-world, such as social networks~\cite{hamilton2017inductive}, financial transaction networks~\cite{wang2020semi} and traffic networks~\cite{yu2017spatio}. 
Graph Neural Networks (GNNs) have shown great ability in modeling graph structured data and are attracting increasing attention~\cite{kipf2016semi,bruna2013spectral,hamilton2017inductive,xu2018powerful}. Generally, GNNs adopt the message-passing process to update node representations by aggregating the information from their neighbors~\cite{velivckovic2017graph,kipf2016semi}. One of the most important and popular tasks that benefits from this message-passing mechanism is node classification in a semi-supervised manner. With this mechanism, labeled nodes can propagate their information to unlabeled nodes~\cite{hamilton2017inductive,tang2020investigating}, thus resulting in superior performance of GNNs.

Despite the great performance of GNNs for semi-supervised node classification, the majority of existing methods assume the training labels are clean; while for many real-world graphs and applications, the collected labels could be noisy and limited. For instance, for the geo-location prediction in social networks, only a small portion of users will fill in the geo-location; and the provided locations can be noisy because users randomly fill in wrong locations to protect their privacy or users have moved to new locations but forget to update them in social networks~\cite{li2012multiple}. Similarly, for  bot detection in social media, the labeling process can be tedious, costly, and error-prone, which can end up with limited noisily labeled nodes~\cite{kudugunta2018deep}.  

The graph with noisy and limited labels could significantly degrade the performance of GNNs for semi-supervised node classification. \textit{First}, recent work has shown that neural networks will overfit to the noisy labels and results in poor generalization performance~\cite{zhang2016understanding,patrini2017making}. As a generalization of neural networks for graphs, GNNs are also likely to have poor performance trained on noisy labels. \textit{Second}, for graphs, the noisy information can propagate through the network topology. Falsely labeled nodes will negatively affect their unlabeled neighbors. Since the graph is sparsely labeled, neighbors of falsely labeled nodes are unlikely to accept the information from nodes with true labels to correct the representations. In addition, many unlabeled nodes will only be able to aggregate information from unlabeled nodes when the labels are limited. 
Thus, the performance of GNNs trained on noisily and sparsely labeled graph would be poor.

Though extensive approaches have been proposed for learning with noisy labels such as loss correction~\cite{patrini2017making,goldberger2016training} and sample selection~\cite{malach2017decoupling,jiang2018mentornet,han2018co,yu2019does,li2020dividemix}, they are not directly applicable for learning GNNs with limited noisy labels. \textit{First}, generally, these methods assume a large amount of noisy labels are available for learning noise distribution or for sampling correct labels. They are challenged by the small label size. \textit{Second}, the majority of  existing work for noisy labels~\cite{patrini2017making,malach2017decoupling,han2018co,li2020dividemix} focus on independent and identically distributed (i.i.d) data such as images, which cannot handle the information propagation of noisy labels on graphs. 
The work on learning a robust GNN with noisy and limited labels is rather limited~\cite{gong2017learning,zhang2020robust}. Therefore, it is important to develop a robust GNN that could deal with noisy and limited labels.

Since the labeled nodes can propagate its information to the unlabeled nodes, it is promising to correct the predictions of unlabeled nodes affected by falsely labeled nodes by linking them with nodes of clean labels. However, in practice, we do not know which labels are clean.  Alternatively, for an unlabeled node $v_i$, we propose to link $v_i$ with labeled nodes of high feature similarity with $v_i$ to make it robust to label noise and facilitate the message passing of GNNs. The basic idea is if two nodes have high feature similarity, they are more likely to have the same label. Thus, if the probability that labeled nodes having correct labels is higher than that of having incorrect labels, by connecting $v_i$ with more labeled nodes of high feature similarity with $v_i$, we can potentially bring more correct label information to $v_i$.  Our theoretical and empirical analysis in Sec~\ref{sec:3.3} verify the effectiveness of linking unlabeled nodes with noisily labeled nodes under mild conditions. 
In addition, with this strategy, we can first train a classifier to obtain accurate pseudo labels to ease the problem of learning with noisy and limited labels. By extending the label set with pseudo labels, more supervision could be utilized to make predictions for unlabeled nodes.  Linking unlabeled nodes with similar nodes of accurate pseudo labels could further reduce the issue of label noise, which is  verified in Sec~\ref{Sec:small_strategy}. Though promising, there are no existing work exploring these strategies for learning GNNs with noisy and limited labels.

Therefore, in this paper, we investigate a novel problem of learning \textit{N}oise-\textit{R}esistant GNNs on sparsely and noisily labeled graphs. In essence, we are faced with two challenges: (i) How to effectively link unlabeled nodes with labeled nodes to alleviate the effects of label noise and benefit the prediction? (ii) Given the graph with noisy and limited labels, how can we obtain accurate pseudo labels? To solve these challenges, we proposed a novel framework named noise-resistant GNN (NRGNN)\footnote{https://github.com/EnyanDai/NRGNN}. NRGNN adopts a GNN-based edge predictor to predict edges
to benefit the classification on graphs with noisy and limited labels. Since the existing edges in the graph generally link nodes in similar attributes~\cite{mcpherson2001birds}, these edges could provide supervision to train a good edge predictor. 
The graph densified by linking unlabeled nodes with similar noisily labeled nodes is utilized to obtain accurate pseudo labels, which extends the label set to provide more supervision for node classification. NRGNN also adopts the edge predictor to link unlabeled with similar extended labeled nodes to further reduce the effects of label noise. In summary, our main contributions are:
\begin{itemize}[leftmargin=*]
    \item We investigate a novel problem of learning noise-resistant GNNs on graphs with noisy and limited labels;
    \item We propose a new framework which can generate accurate pseudo labels and assign high-quality edges between unlabeled nodes and (pseudo) labeled nodes to alleviate label noise issue;
    \item Theoretical and empirical analysis are conducted to verify the effectiveness of the proposed strategies against label noise;
    \item Extensive experiments on real-world datasets demonstrate the effectiveness of the proposed NRGNN in node classification on graphs with noisy and limited labels.
\end{itemize}
\section{Related Work}
In this section, we present the related literature of graph neural networks and deep learning with noisy labels.

\subsection{Graph Neural Networks}
Graph neural networks (GNNs) have shown great ability in modeling graph structured data. They have achieved remarkable success in various applications such as social networks~\cite{hamilton2017inductive,dai2021say}, financial transaction networks~\cite{wang2020semi} and traffic networks~\cite{yu2017spatio,zhao2020semi}. Based on the definition of graph convolution, GNNs can be generally divided into two categories, i.e., spectral-based~\cite{bruna2013spectral,henaff2015deep,defferrard2016convolutional,kipf2016semi,levie2018cayleynets} and spatial-based~\cite{velivckovic2017graph,hamilton2017inductive,chen2018fastgcn,ying2018graph,zhao2021graphsmote,tang2020transferring}. \citeauthor{bruna2013spectral} \cite{bruna2013spectral} first explored spectral-based GNNs by utilizing the spectral filter on the local spectral space. Since then, various spectral-based methods are developed for further improvements \cite{henaff2015deep,defferrard2016convolutional,kipf2016semi,levie2018cayleynets}. For instance, \citeauthor{kipf2016semi}~\cite{kipf2016semi} propose Graph Convolutional Network (GCN) which simplifies the graph convolution. Spatial-based graph convolution directly updates the node representation by aggregating its neighborhoods' representations \cite{niepert2016learning,gilmer2017neural,hamilton2017inductive,ying2018graph}. 
For example, graph attention network (GAT)~\cite{velivckovic2017graph} applies the self-attention mechanism into the aggregation of spatial graph convolution. Graph Isomorphism Network (GIN)~\cite{xu2018powerful} is proposed to learn more powerful representations of the graph structures.
Moreover, various spatial-based methods are investigated to solve the scalability issue of GNNs~\cite{hamilton2017inductive, chen2018fastgcn}.

However, as a generalization of neural networks on graph structured data, GNNs are also vulnerable to noisy labels~\cite{zhangadversarial,nt2019learning}. In addition, due to the message passing mechanism of GNNs, the noisy label information will pass to the unlabeled nodes, which severely degrades the performances of GNNs. For example, \cite{nt2019learning} shows that the performance of GNNs will drop significantly when noises are added to the training labels. However, very few efforts are taken to address the problem of learning GNNs on graphs with noisy labels~\cite{nt2019learning,patrini2017making}.  D-GNN~\cite{nt2019learning} applied the backward loss correction~\cite{patrini2017making} to reduce the effects of noisy labels. \citeauthor{zhangadversarial}~\cite{zhangadversarial} avoid the overfitting of the noisy labels by adding a regularization which encourages the learned representations well predict the community labels. Our proposed framework is inherently different from aforementioned methods. We investigate a novel framework which could achieve robustness towards noisy labels in graphs by carefully connecting unlabeled nodes with (pseudoly) labeled nodes.

\subsection{Deep Learning with Noisy Labels} 
It is shown in~\cite{zhang2016understanding} that a standard deep neural network will overfit to the noisy labels and results in poor generalization performance. Extensive studies have been investigated to address this problem on i.i.d data such as images, which can be generally categorized into two groups: loss correction~\cite{patrini2017making,goldberger2016training,reed2014training, ma2018dimensionality} and sample selection~\cite{malach2017decoupling,jiang2018mentornet,han2018co,yu2019does,li2020dividemix}. The loss correction methods correct the loss of training samples with noisy labels. For example, \citeauthor{goldberger2016training}~\cite{goldberger2016training} propose a noise adaptation layer to automatically learn the noise transition matrix and sequentially apply it to correct the loss in S-model. \citeauthor{patrini2017making}~\cite{patrini2017making} estimate the label corruption matrix and propose two ways of correcting the loss, i.e., forward and backward correction. Bootstrap~\cite{reed2014training} handles noisy labels by augmenting the prediction objective with a notion of consistency. 
The sample selection methods aim to find the clean samples during the training process. For example,  Decoupling~\cite{malach2017decoupling} deploys two networks to select clean samples and update the two networks with the clean samples obtained from each other.
MentorNet~\cite{jiang2018mentornet} pre-trains a teacher network to reweight the samples during the training process of the student network. Coteaching~\cite{han2018co} also employs two networks and selects the small-loss samples as clean samples for each other. Moreover, Coteaching+\cite{yu2019does} incorporates additional rule of updating when disagreement to improve the performance of Coteaching. Recently, methods that utilize the data points that are not selected as clean samples by semi-supervised learning methods are also investigated~\cite{nguyen2019self, li2020dividemix}. 

However, the aforementioned approaches are dedicated to i.i.d data, which may not be directly applicable to GNNs for handing noisy labels because the noisy information can propagate via message passing of GNNs. Therefore, we propose a novel approach NRGNN to handle the label corruption on the graph-structured data. Furthermore, we address the challenge of learning with labels that are often noisy and limited in graphs.
\section{Preliminaries}
In this section, we firstly introduce the basic design of GNNs. Next, two strategies of addressing the problem of learning on noisily and sparsely labeled graphs are analyzed theoretically and empirically.
\label{Sec:3}
\subsection{Notation}
\label{Sec:notation}
We use $\mathcal{G}=(\mathcal{V},\mathcal{E})$ to denote a graph, where $\mathcal{V}=\{v_1,...,v_N\}$ is the set of $N$ nodes, $\mathcal{E} \subseteq \mathcal{V} \times \mathcal{V}$ is the set of edges, and $\mathbf{A} \in \mathbb{R}^{N \times N}$ is the adjacency matrix of the graph $\mathcal{G}$, where $\mathbf{A}_{ij}=1$ if nodes ${v}_i$ and ${v}_j$ are connected, otherwise $\mathbf{A}_{ij}=0$. $\mathbf{X}=\{\mathbf{x}_1,...,\mathbf{x}_N\}$ is the set of node attributes with $\mathbf{x}_i$ being the node attributes of node $v_i$. $\mathcal{V}_L=\{v_1,...,v_l\}$ is a set of labeled nodes. $\mathcal{V}_U = \mathcal{V}-\mathcal{V}_L$ is a set of unlabeled nodes. The provided labels of $\mathcal{V}_L$ are corrupted by noise, which are denoted as $\mathcal{Y}_N=\{y_1^n,...,y_l^n\}$. And $\mathcal{Y}_T=\{y_1^t,...,y_l^t\}$ is used to represent the true labels. 


\subsection{Preliminaries about GNN}
Graph neural networks (GNNs) utilize the node features and the graph structures to learn presentations for prediction. Specifically, each layer of GNNs will update the representations of the nodes using the representations of the neighborhood nodes. Thus, the representations after $k$ layers' aggregation would capture the information of the $k$-hop network neighborhoods, which would benefit the node classification. Generally, the updating process of the $k$-th layer in GNN is formally stated as:
\begin{equation}
\begin{aligned}
    \mathbf{a}^{(k)}_v & = \text{AGGREGATE}^{(k-1)}(\{\mathbf{h}^{(k-1)}_u: u \in \mathcal{N}(v)\}),
    \label{eq:GNN_a} \\
    \mathbf{h}^{(k)}_{v} & =\text{COMBINE}^{(k)}(\mathbf{h}^{(k-1)}_v, \mathbf{a}^{(k)}),
\end{aligned}
\end{equation}
where $\mathbf{h}^{(k)}_v$ is the representation vector of the node $v \in \mathcal{V}$ at $k$-th layer and $\mathcal{N}(v)$ is  a set of neighborhoods of $v$. GCN is one of the most popular GNN structures, which could be viewed as a special case of Eq.(\ref{eq:GNN_a}). Each layer of GCN can be written as:
\begin{equation}
    \mathbf{H}^{(k+1)} = \sigma(\tilde{\mathbf{A}}\mathbf{H}^{(k)}\mathbf{W}^{(k)}),
\end{equation}
where $\mathbf{H}^{(k)}$ is the representation matrix of the output of the $k$-th layer; $\tilde{\mathbf{A}}=\mathbf{D}^{-\frac{1}{2}}(\mathbf{A}+\mathbf{I})\mathbf{D}^{-\frac{1}{2}}$ is the normalized adjacency matrix and $\mathbf{D}$ is a diagonal matrix with $D_{ii}=\sum_{i}A_{ij}$. $\mathbf{I}$ is the identity matrix and $\sigma$ is an activation function such as ReLU.

\subsection{Problem Definition}
Given the notation in Sec~\ref{Sec:notation}, the problem of learning a robust GNN with noisy and limited labels is formally defined as:
\newtheorem{problem}{Problem}
\begin{problem}
Given a graph $\mathcal{G}=(\mathcal{V}, \mathcal{E}, \mathbf{X})$ with a small set of nodes $\mathcal{V}_L \in \mathcal{V}$ provided with noisy labels $\mathcal{Y}_N$, we aim to learn a robust GNN which predicts the true labels of the unlabeled nodes, i.e.,
\begin{equation}
    f(\mathcal{G}, \mathcal{Y}_N) \rightarrow \mathcal{\hat{Y}}_U
\end{equation}
where $f$ is the function we aim to learn and $\hat{\mathcal{Y}}_U$ is the set of predicted labels for unlabeled nodes. 
\end{problem}

\subsection{How the Size of Noisily Labeled Neighbors Affect the Node Classification}
\label{sec:3.3}

For a trained $K$-layer GNN with a set of learned parameters $\theta=\{\mathbf{W}^{(1)},...,\mathbf{W}^{(K)}\}$, it makes predictions by $\mathbf{Y}=\tilde{\mathbf{A}}\mathbf{H}^{(K)}\mathbf{W}^{(K)}$. Since the parameters $\theta$ are well trained for node classification, the $K$-th latent representations, i.e., $\mathbf{S}=\mathbf{H}^{(k)}\mathbf{W}^{(K)}$ could be treated as predictions of the nodes~\cite{dong2020equivalence}. And the final predictions are obtained by the aggregation of $\mathbf{S}$, i.e., $\mathbf{Y}=\tilde{\mathbf{A}}\mathbf{S}$. Let's treat $s_{ik}$ as the predicted probability that node $v_i$ belongs to class $k$. For an unlabeled $v_u \in \mathcal{V}_U$ belonging to class $c$, we consider three types of neighbors: (i) an unlabeled node $v_a \in \mathcal{V}_U$; (ii) a node $v_b \in \mathcal{V}_L$ labeled as $c$; and (iii) a node $v_d \in \mathcal{V}_L$ labeled to a class other than $c$.
Since the GCN is optimized to make $s_{bc}$ close to 1 and $s_{dc}$ close to 0, generally we could have $\mathbb{E}(s_{bc}) > \mathbb{E}(s_{ac}) > \mathbb{E}(s_{dc})$. To simplify the analysis, we assume that nodes with high feature similarity belong to the same class.
Then, we can have the following theorem which indicates that linking an unlabeled node with similar labeled nodes could increase the robustness against label noise.   

\begin{theorem}
We consider an unlabeled node $v_u \in \mathcal{V}_U$ which belongs to class $c$. It is linked with $n$ unlabeled nodes and $m$ labeled nodes. The ratio of intra-class edges is $h$. 
Assume that:
\begin{enumerate}
    \item For labeled nodes, a node belonging to class $c$ is more likely to be labeled as $c$ than a node not belonging to class $c$;
    \item The probability $p_t$ that a node belonging to class $c$ is labeled as $c$ meets this constraint: $p_t > \frac{\mathbb{E}(s_{ac})-\mathbb{E}(s_{dc})}{\mathbb{E}(s_{bc})-\mathbb{E}(s_{dc})}$.
\end{enumerate}
Then, linking $v_u$ with more similar noisily labeled nodes $v_l \in \mathcal{V}_L$ can on average improve its predicted probability of belonging to class $c$, i.e., improve $y_{uc}=\frac{1}{d_u}\sum_{j \in \mathcal{N}(v_i)} s_{jc}$. 
\label{theorem:labeled}
\end{theorem}
The proof of this theorem is presented in Appendix~\ref{app:proof_labeled}.
When a graph network network for multi-class classification is corrupted with label noise, the predicted probability of an unlabeled node $v_a \in \mathcal{V}_U$ would be much smaller that the probability that a node labeled as $c$. Therefore, the assumption that $p_t > \frac{\mathbb{E}(s_{ac})-\mathbb{E}(s_{dc})}{\mathbb{E}(s_{bc})-\mathbb{E}(s_{dc})}$ could be generally satisfied.

\noindent \textbf{Empirical analysis}: 
According to the Theorem~\ref{theorem:labeled}, if we could link the unlabeled nodes with more similar labeled nodes belonging to the same class, we will have a more robust model. On the contrary, linking unlabeled nodes may not be useful. To empirically verify this, we utilize the cosine similarity scores of the raw features to identify similar nodes. Then, edges could be added based on the similarity scores.
More specifically, we compare the results of the following methods:
\begin{itemize}[leftmargin=*]
    \item \textbf{Initial $\mathcal{G}$}: We train a GCN on the initial graph structure with noisy labels as the baseline.
    \item \textbf{Link $\mathcal{V}_L$}: For $v_u \in \mathcal{V}_U$ and $v_l \in \mathbf{V}_L$, if their raw feature cosine similarity is larger than $t$, we add a link between them. Then, a GCN trained on $\mathcal{G}$ will make predictions with the new graph.
    \item \textbf{Link $\mathcal{V}_U$}: Unlabeled nodes will be linked with other unlabeled nodes if they have high cosine similarity of features. Similarly, a GCN trained on $\mathcal{G}$ will make predictions with the new graph.
\end{itemize}
We conduct experiments on widely used benchmark Cora and Citeseer~\cite{sen2008collective}. In both datasets, we randomly sample 5\% nodes as labeled nodes. And labels are corrupted by randomly flipping the true labels to other class with a probability of $p$. More specifically, we vary the noise rate. i.e. the probability that given labels is wrong, from 10\% to 30\% with a step of 10\%. The thresholds of cosine similarity are selected based on the validation set. We report average results of 5 runs in Table~\ref{tab:pre}. We could have the following observations:
\begin{itemize}[leftmargin=*]
    \item Linking the unlabeled nodes with unlabeled nodes shows no difference from the results of training on initial graph.
    \item Even a simple strategy based on raw feature cosine similarity to link unlabeled nodes with labeled nodes could benefit node classification trained on noisy labels significantly. 
    \item When the noise rate is raised to 0.3, linking unlabeled nodes with similar labeled nodes still shows its effectiveness.
\end{itemize}


\subsection{A Strategy On Graphs with Small Amount of Noisy Labels}
\label{Sec:small_strategy}
With the analysis in Sec.~\ref{sec:3.3}, we find that linking more existing noisily labeled nodes with the unlabeled nodes could make more robust predictions. However, the size of noisily labeled nodes are often very small in graph-structure data. And an unlabeled node may have small node similarity with the labeled nodes. In this situation, the benefits from the strategy described in Sec~\ref{sec:3.3},
would be largely limited. A strategy to address this problem is to obtain accurate pseudo labels $\mathcal{V}_P$. As a result, we could have an extended label set $\mathcal{V}_A = \mathcal{V}_L \cup \mathcal{V}_P$. Then an unlabeled node can have more similar nodes in $\mathcal{V}_A$ to have a more robust model. In addition, more supervision from pseudo labels can be utilized. Let $s_{pc}$ denotes the predicted probability that node $v_p \in \mathcal{V}_P$ belongs to class $c$
based on the $K$-th latent representations, i.e., $\mathbf{S} = \mathbf{H}^{(k)}\mathbf{W}^{(K)}$.
The following theorem verifies the effectiveness of this strategy when $s_{pc}$ meets a mild constraint.
\begin{table}[t]
    \small
    \centering
    \caption{Accuracy(\%) of node classification with noisy labels.}
    \vskip -1.5em
    \begin{tabularx}{0.95 \linewidth}{|p{0.095\linewidth}|p{0.14\linewidth}|XXX|}
    \hline
    Dataset & Noise Rate & Initial $\mathcal{G}$ & Link $\mathcal{V}_U$  & Link $\mathcal{V}_L$ \\
    \hline
    \multirow{3}{*}{Cora} 
    & 0.1 & 77.9 $\pm 0.3$ & 77.8 $\pm 0.5$ & \textbf{78.7} $\pm \textbf{0.4}$ \\
    & 0.2 & 72.8 $\pm 1.8$ & 72.8 $\pm 1.0$ & \textbf{74.0} $\pm \textbf{0.9}$ \\
    & 0.3 & 65.6 $\pm 0.8$ & 65.8 $\pm 1.7$ & \textbf{68.5} $\pm \textbf{1.4}$ \\
    \hline
    \multirow{3}{*}{Citeseer} 
    & 0.1 & 68.1 $\pm 0.8$& 68.1 $\pm 0.6$  &\textbf{69.0} $\pm \textbf{1.0}$ \\
    & 0.2& 64.9 $\pm 1.7$ & 65.2 $\pm 0.8$  &\textbf{66.4} $\pm \textbf{1.5}$ \\
    & 0.3& 60.4 $\pm 2.5$ & 61.8 $\pm 1.0$  &\textbf{62.7} $\pm \textbf{1.0}$\\
    \hline
    \end{tabularx}
    \label{tab:pre}
    \vskip -1em
\end{table}
\begin{table}[t]
    \small
    \centering
    \caption{Accuracy(\%) of node classification with noisy labels.}
    \vskip -1.5em
    \begin{tabularx}{0.98 \linewidth}{p{0.095\linewidth}XX>{\centering\arraybackslash}p{0.24\linewidth}X}
    \toprule
    Dataset & Initial $\mathcal{G}$ & Link $\mathcal{V}_L$   & Link $\mathcal{V}_L$ (Retrain) & Link $\mathcal{V}_A$ \\
    \midrule
    Cora & 72.8 $\pm 1.8$ & 74.0 $\pm 0.9$ &75.4 $\pm 1.0$ & \textbf{77.1} $\pm \textbf{1.3}$\\
    Citeseer & 64.9 $\pm 1.7$ & 66.4 $\pm 1.5$  &66.5 $\pm 1.5$ & \textbf{68.0} $\pm \textbf{1.4}$\\
    \bottomrule
    \end{tabularx}
    \label{tab:pre_2}
    \vskip -1.5em
\end{table}
\begin{theorem}
We consider an unlabeled node $v_u \in \mathcal{V}_U$ which belongs to class $c$. It is linked with $n$ unlabeled nodes and $m$ labeled nodes. Let $p$ denotes the probability that the existing linked labeled nodes is labeled as $c$.
For a node $v_p \in \mathcal{V}_P$ which is provided with pseudo label, if $\mathbb{E}(s_{pc}) > max(\mathbb{E}(s_{ac}),p\mathbb{E}(s_{bc})+(1-p)\mathbb{E}(s_{dc}))$, then, linking $v_u$ with $v_p$ can improve its predicted probability of belonging to class $c$, i.e., $y_{uc}=\frac{1}{d_u}\sum_{j \in \mathcal{N}(v_i)} s_{jc}$. 
\label{theorem:pseudo}
\end{theorem}
The details of the proof is listed in Appendix~\ref{app:proof_pseudo}.
To obtain pseudo labels which meet the assumption to benefit the predictions under label noise, we could utilize the strategy described in Sec~\ref{sec:3.3} to give better predictions. Furthermore, we can select the predictions whose confidence scores are high.

\noindent \textbf{Empirical analysis}: To show the effectiveness of the strategy of utilizing accurate pseudo labels, we conduct experiments with the following process: 1) obtain a GNN classifier using the strategy of linking unlabeled nodes and labeled nodes based on cosine similarity; 2) select the predictions of unlabeled nodes whose confidence scores are high as pseudo labels $\mathcal{V}_P$ to compose extended label set $\mathcal{V}_A=\mathcal{V}_L \cup \mathcal{V}_P$; 3) link $\mathcal{V}_U$ and $\mathcal{V}_A$ based on cosine similarity of raw features, and train a GCN with the new graph with accurate pseudo labels and noisy labels. This process is named as \textbf{Link $\mathcal{V}_A$}. And the results of 5 runs on Cora and Citeseer with 20\% uniform noise are presented in Table~\ref{tab:pre_2}. To make a fair comparison, we also retrain the GCN on the graph densified by linking unlabeled nodes with similar labeled nodes.
From the Table~\ref{tab:pre}, we could find that with the strategy of utilizing accurate 
pseudo labels, the model become more robust to the label noise.

\begin{figure}
    \centering
    \includegraphics[width=0.95\linewidth]{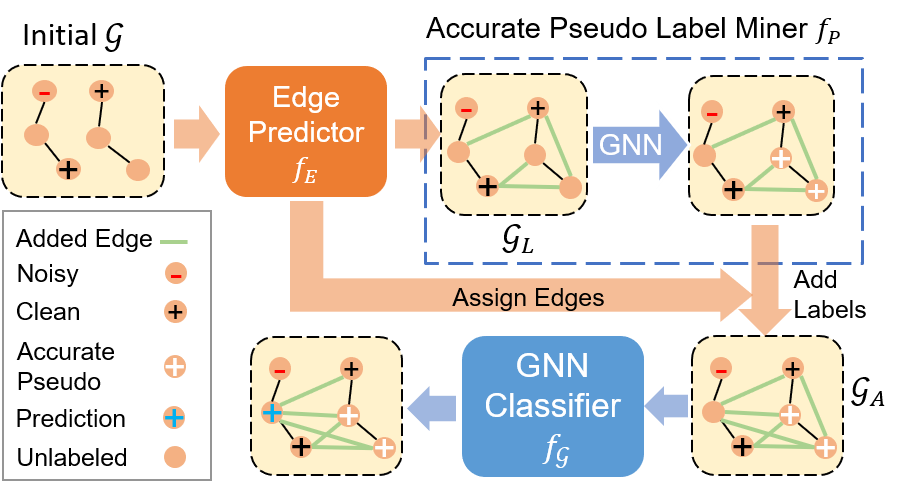}
    \vskip -1.5em
    \caption{The overall framework of our method.}
    \label{fig:framework}
    \vskip -1.8em
\end{figure}

\section{Methodology}
In this section, we present the details of the proposed framework NRGNN. As shown in Sec~\ref{Sec:3}, carefully linking unlabeled nodes with nodes with noisy labels or accurate pseudo labels could benefit the learning of GNNs on noisily and sparsely labeled graphs. 
However, there are two main challenges: (i) How to accurately add edges between unlabeled nodes and extended labeled nodes of the same class to benefit the prediction? and  (ii) Given the graph with limited noisy labels, how to obtain accurate pseudo labels?  To solve these two challenges, we propose to learn a GNN-based edge predictor using node attributes to assign high-quality edges. To obtain more high-quality pseudo labels, we utilize the graph densified by linking similar unlabeled nodes and noisily labeled nodes with the GNN-based edge predictor.
An illustration of the proposed framework is shown in Fig.~\ref{fig:framework}, which is composed of an edge predictor $f_E$, a pseudo label miner $f_P$, and a GNN classifier $f_{\mathcal{G}}$. The edge predictor takes the initial graph $\mathcal{G}$ as input to predict edges. $f_E$ will first link similar nodes between $\mathcal{V}_U$ and $\mathcal{V}_L$ to obtain $\mathcal{G}_L$, which could benefit the pseudo label miner. The pseudo label miner $f_P$ adopts a GNN classifier trained on $\mathcal{G}_L$ to collect nodes with high-confident pseudo labels, denoted as $\mathcal{V}_P$. With the extended label set, the edge predictor $f_E$ further connects unlabeled nodes $V_U$ with similar nodes in $\mathcal{V}_L\cup \mathcal{V}_P$ to form a new graph $\mathcal{G}_A$, which helps to propagate the information from $\mathcal{V}_L\cup \mathcal{V}_P$ to unlabeled nodes. The final GNN classifier $f_\mathcal{G}$ takes $\mathcal{G}_A$ with the label set $\mathcal{V}_L\cup \mathcal{V}_P$ for robust prediction. Next, we introduce each component in detail.

\subsection{Edge Prediction}
In many real-world networks, linked nodes generally have the same labels or similar features~\cite{mcpherson2001birds}. For instance, papers are more likely to cite papers belonging to the same research field, and friends tend to share similar interests~\cite{newman2018networks}. In addition, many real-world graphs are very sparse and contains lots of missing links. For example, a social media user may miss a lot of potential friends sharing same interests and only follow a small number of people due to time limitation in exploring friends online. Thus, link prediction algorithms can learn from the attributed graph to predict the missing links, which provides one direction for us to link nodes.

Our preliminary analysis in Sec.~\ref{Sec:3} has shown that by simply using the node feature similarity based link prediction to connect unlabeled nodes with labeled nodes could help to improve the performance of GNNs with noisy labels. However, the simple approach only considers node features to measure the similarity. While for graphs, the local graph structure also provides another perspective for measuring similarities. To better predict the missing links, we propose to use a GNN-based edge predictor. Instead of simply relying on node features, the GNN-based edge predictor learns node representations capturing both node features and local graph structure, and predict links based on the learned representations, which could improve the link prediction performance.  
Following~\cite{kipf2016variational}, our edge predictor adopts the GCN to learn node representations as:
\begin{equation}
    \mathbf{Z} = GCN(\mathbf{A},\mathbf{X}).
\end{equation}
Let $\mathbf{z}_i$ and $\mathbf{z}_j$ denote the representations of node $v_i$ and $v_j$, respectively. The closer $\mathbf{z}_i$ and $\mathbf{z}_j$ are, the more likely $v_i$ and $v_j$ are linked. Thus, the probability that $v_i$ and $v_j$ are linked can be calculated as
\begin{equation}
    \mathbf{S}_{ij}=\sigma(\mathbf{z}_i\mathbf{z}_j^T),
\end{equation}
where $\sigma(\cdot)$ is the activation function. Because the learned weights $\mathbf{S}$ would be fed into other modules and trained end-to-end, we use ReLU as the activation function to avoid the gradient vanishing~\cite{he2017neural}. 

If the edge predictor could well reconstruct the adjacency matrix $\mathbf{A}$, then it would be good at predicting missing links. Thus, following existing work~\cite{kipf2016variational}, we use the adjacency matrix reconstruction as the loss.
However, the majority of the elements in $\mathbf{A}$ are 0's, which could dominate the loss function and result in edge predictor $f_E$ simply outputting 0's.  To avoid this, we apply negative sampling~\cite{mikolov2013distributed}, i.e., for each positive sample $A_{ij}=1$, we randomly sample $K$ nodes which are not connected with node $j$ as negative samples. 
With the negative sampling, the loss function could be formally written as:
\begin{equation}
\small
    \min_{\theta_{E}}\mathcal{L}_E = \sum_{v_i \in \mathcal{V}}\sum_{v_j \in \mathcal{N}(v_i)} \Big((\mathbf{S}_{ij}-1)^2 + \sum_{n=1}^K \mathbb{E}_{v_n \sim P_n(v_i)} (\mathbf{S}_{in}-0)^2\Big)
    \label{eq:edge}
\end{equation}
where $\theta_{E}$ is the set of parameters of $f_E$, $\mathcal{N}(v_i)$ represents the neighbors of node $v_i$, and $P_n(v_i)$ is the distribution of the nodes which have no connections with node $v_i$ in the graph. With the GNN-based edge predictor trained with Eq.(\ref{eq:edge}), we could predict useful missing edges to link unlabeled nodes and labeled nodes to benefit the robust classification with noisy and limited labels.

\subsection{Accurate Pseudo Label Prediction}
According to the analysis in Sec.~\ref{Sec:small_strategy}, more accurate pseudo labels would better facilitate the training of GNNs with noisy and limited labels. Thus, in this subsection, we describe how to obtain accurate pseudo labels. Since connecting unlabeled nodes with labeled nodes by link prediction can help improve the node classification of a GCN on graph with noisy labels, we propose to first predict the missing edges between $\mathcal{V}_U$ and $\mathcal{V}_L$ with the edge predictor $f_E$. Then, we could obtain a densified graph $\mathcal{G}_L$ to train a more accurate pseudo label miner. Specifically, for $v_i \in \mathcal{V}_U$ and $v_j \in \mathcal{V}_L$, if $S_{ij}$ is larger than a threshold $t$, then $v_i$ and $v_j$ are more likely to have the same label and we would connect them. If $S_{ij} < t$, then the probability that $v_i$ and $v_j$ having the same label is small and we don't want to include such links. Thus, the process of obtaining the adjacency matrix of $\mathcal{G}_L$ could be formally stated as:
\begin{equation}
    \mathbf{S}^L_{ij} = \left\{ \begin{array}{ll}
    1 & \mbox{if $v_j \in \mathcal{N}(v_i)$} ;\\
     \mathbf{S}_{ij} & \mbox{else if $\mathbf{S}_{ij} > t$, $v_i \in \mathcal{V}_U$ and $v_j \in \mathcal{V}_L$} ;\\
    0 & \mbox{else},\end{array} \right.
    \label{eq:generate_graph}
\end{equation}
where $\mathbf{S}^L_{ij}$ indicates the weight of edges between node $v_i$ and $v_j$ in $\mathcal{G}_L$, and $t$ is the threshold to filter out edges with small weights.  With $\mathbf{S}^L$, we can train a GNN classifier as pseudo label miner $f_P$. The pseudo labels of nodes $\mathcal{V}$ is predicted as:
\begin{equation}
    \mathbf{\hat{Y}}^P = GNN(\mathbf{S}^L, \mathbf{X}),
\end{equation}
where GNN is flexible to various models such as GCN~\cite{kipf2016semi} and GIN~\cite{xu2018powerful}. Its training objective function can be written as:
\begin{equation}
    \min_{\theta_{P}}\mathcal{L}_P = \sum_{v_i\in\mathcal{V}_L} l(\hat{y}_i^P,y_i),
    \label{eq:loss_pseudo}
\end{equation}
where $\theta_P$ is the parameters of the pseudo label miner $f_P$, $\hat{y}_i^P$ is the prediction of node $v_i$ from $f_P$, and $l(\cdot)$ is the cross entropy loss. With $\mathbf{S}^L$, we can reduce the negative effects of label noise and have more reliable pseudo labels for the unlabeled nodes. Intuitively, the pseudo label whose confidence score is high should be more likely to be correct.
Let $\hat{y}_{ic}^P$ denotes the predicted probability that node $v_i$ belongs to the class $c$.
Then the accurate pseudo labels is obtained by the following process:
\begin{equation}
    \mathcal{Y}_P=\{\hat{y}_i^P \in \mathcal{\hat{Y}}_U^P; \hat{y}_{ic}^P > T_p\},
    \label{eq:mine_pseudo}
\end{equation}
where $\mathcal{\hat{Y}}_U^P$ is the set of predictions from the pseudo label miner for unlabeled nodes, and $T_p$ is the threshold to select the accurate pseudo labels.

\subsection{Robust Classification with Edge Predictor and Accurate Pseudo Labels}
The accurate pseudo labels $\mathcal{Y}_P$ could facilitate the classification with noisy and limited labels in two folds: (i) accurate pseudo labels could provide more supervision for node classification; and (ii) edges linking unlabeled nodes and accurate pseudo labeled nodes could be added to reduce the effects of label noise. To fully utilize the pseudo labels, we adopt the edge predictor $f_E$ to assign missing links between unlabeled nodes $\mathcal{V}_U$ and extended labeled nodes $\mathcal{V}_A = \mathcal{V}_L \cup \mathcal{V}_P$, where $\mathcal{V}_P$ is the node set with accurate pseudo labels $\mathcal{Y}_P$. Similar to the construction of $\mathbf{S}^L$, we use the same threshold $t$ to select links. 
This process is written as:
\begin{equation}
    \mathbf{S}^A_{ij} = \left\{ \begin{array}{ll}
    1 & \mbox{if $v_j \in \mathcal{N}(v_i)$} ;\\
     \mathbf{S}_{ij} & \mbox{else if $\mathbf{S}_{ij} > t$, $v_i \in \mathcal{V}_U$ and $v_j \in \mathcal{V}_A$} ;\\
    0 & \mbox{else},\end{array} \right.
    \label{eq:generate_graph_all}
\end{equation}
where $\mathbf{S}^A_{ij}$ denotes the weight of edge linking node $v_i$ and $v_j$. With the extended label set $\mathcal{V}_A$ providing more label information, and the new adjacency matrix facilitating the information propagation from $\mathcal{V}_A$ to $\mathcal{V}_U$, we can train a more robust GNN classifier against the noisy for label prediction as
\begin{equation}
    \hat{\mathbf{Y}} = f_\mathcal{G} (\mathbf{S}^A, \mathbf{X}) 
\end{equation}
where $\hat{\mathbf{Y}}$ is the final label prediction. Similar to the accurate pseudo label miner, the GNN classifier $f_\mathcal{G}$ is flexible to various GNNs such as GCN~\cite{kipf2016semi} and GIN~\cite{xu2018powerful}. The training of $f_{\mathcal{G}}$ utilizes the supervision from both noisy labels and accurate pseudo labels. The loss function can be written as:
\begin{equation}
    \mathcal{L}_\mathcal{G} = \sum_{v_i \in \mathcal{Y}_A} l(\hat{y}_i, y_i),
    \label{eq:loss_G}
\end{equation}
where $y_i$ denotes the noisy label or accurate pseudo label of the node $v_i \in \mathcal{V}_A$ and $\hat{y}_i$ denotes the prediction of node $v_i \in \mathcal{V}_A$.

\subsection{Final Objective Function}
With edge predictor adding links for facilitating the information propagation, pseudo label miner providing more labels and the GNN classifier predicting the labels, the overall loss function can be written as:
\begin{equation}
    \argmin_{\theta_E,\theta_P, \theta_\mathcal{G}} \mathcal{L}_\mathcal{G} +  \alpha \mathcal{L}_E + \beta \mathcal{L}_P,
    \label{eq:loss_final}
\end{equation}
where $\theta_E$, $\theta_P$, and $\theta_\mathcal{G}$ are the parameters of edge predictor $f_E$, accurate label miner $f_P$ and GNN classifier $f_\mathcal{G}$, respectively. $\alpha$ and $\beta$ are hyperparameters to balance the contributions of adjacency matrix reconstruction loss of $f_E$ and the loss of pseudo label miner. $f_{\mathcal{G}}$, $f_E$ and $f_P$ are jointly trained together with Eq.(\ref{eq:loss_final}). The details of the \textit{training algorithm} is presented in Appendix~\ref{app:alg}.

\section{experiments}
\label{Sec:experiments}
In this section, we conduct experiments on real-world datasets to show the effectiveness of the proposed framework. In particular, we aim to answer the following research questions:
\begin{itemize}[leftmargin=*]
    \item \textbf{RQ1} Is the proposed framework NRGNN robust to different types and levels of label noise?
    \item \textbf{RQ2} Is the proposed framework effective under different sizes of noisy labels and graph sparsity?
    \item \textbf{RQ3} Is NRGNN flexible to various GNN backbones and how do the edge predictor and pseudo label miner contribute to NRGNN?
\end{itemize}

\subsection{Experimental Settings}

\label{Sec:ex_settings}
\subsubsection{Datasets}
\label{Sec:dataset}
We conduct experiments on four widely used benchmark datasets, i.e., Cora, Citeseer, Pubmed~\cite{sen2008collective} and DBLP~\cite{pan2016tri}. The statistics of the datasets are presented in Table~\ref{tab:dataset} in Appendix. The validation and test sets are kept the same as the cited papers to keep consistency. As for the training set, we randomly sample 5\% nodes for Cora and Citeseer. For large datasets, i.e., Pubmed and DBLP, we sample 1\% nodes to compose the training set. All the training set has no overlap with validation and test sets. Since  the labels of these datasets are clean, following~\cite{reed2014training,patrini2017making,yu2019does}, we corrupt the labels of training and validation set with two types of label noises:
\begin{itemize}[leftmargin=*]
    \item \textbf{Uniform Noise}: 
    The labels have a probability of $p$ to be uniformly flipped to other classes. 
    \item \textbf{Pair Noise}: Labelers are assumed to make mistakes only within the most similar pair classes. More specifically, labels have a probability of $p$ to flip to their pair class.
\end{itemize}

\subsubsection{Implementation Details} 
\label{Sec:implement}
We report the average results with standard deviations of 5 runs for all experiments. A two-layer GCN whose hidden dimension is 16 is deployed as the backbone of the edge predictor. Similarly, the pseudo label miner and GNN classifier also uses two-layer GCNs as backbones, respectively. Note that our framework is flexible to use various GNNs, which is demonstrated by the experimental results in Sec~\ref{Sec:abl}. All hyper-parameters are tuned based on the validation set. We vary $\alpha$ and $\beta$ among $\{0.001, 0.01, 0.1, 1, 10\}$ and $\{0.001,0.01,0.1,1,10, 100\}$, respectively. As for $t$ and $T_p$, we fix them as 0.1 and $0.8$ for all the datasets. And the number of negative samples $K$ is set as 50.

\begin{table*}[t]
    \small
    \centering
    \caption{Node classification performance (Accuracy (\%)$\pm$Std) under various types of noise.}
    \vskip -1.5em
    \begin{tabularx}{0.95\textwidth}{|p{0.05\textwidth}|p{0.06\textwidth}|CC>{\centering\arraybackslash}p{0.085\linewidth}CCCCC|}
    \hline
    Dataset & Noise & GCN & GIN & Self-Training & Forward & Coteaching+ & D-GNN & CP  & Ours \\
    \hline
    
    \multirow{2}{*}{Cora} & Uniform & 72.8 $\pm 1.8$  & 72.3 $\pm 0.9$ & 75.6 $\pm 1.8$ & 73.7 $\pm 0.7$ & 73.6 $\pm 1.7$ & 72.4 $\pm 1.8$ & 74.8 $\pm 1.3$ & \textbf{80.4} $\pm \textbf{0.5}$\\
    & Pair & 74.1 $\pm 0.7$ & 74.7 $\pm 1.4$ & 76.4 $\pm 1.4$ & 76.0 $\pm 0.7$ & 73.8 $\pm 1.4$ & 73.5 $\pm 1.6$ & 75.2 $\pm 1.4$ & \textbf{79.5} $\pm \textbf{0.4}$ \\
    \hline
    \multirow{2}{*}{Citeseer} & Uniform & 64.9 $\pm 1.7$ & 65.7 $\pm 2.1$ & 67.8 $\pm 1.4$ & 65.0 $\pm 1.5$ & 66.4 $\pm 1.3$ & 64.9 $\pm 1.3$ & 66.0 $\pm 1.6$  & \textbf{70.1} $\pm \textbf{1.3}$\\
    & Pair & 60.3 $\pm 1.0$ & 61.6 $\pm 1.0$ & 62.0 $\pm 1.6$ & 61.6 $\pm 0.4$ & 65.1 $\pm 2.1$ & 62.3 $\pm 1.2$ & 62.0 $\pm 1.0$  & \textbf{67.8} $\pm \textbf{1.3}$\\
    \hline
    \multirow{2}{*}{Pubmed} & Uniform & 77.3 $\pm 0.9$ & 77.4 $\pm 0.5$ & 78.2 $\pm 0.4$ & 77.5 $\pm 0.4$ & 78.6 $\pm 0.4$ & 77.6 $\pm 0.3$ & 78.6 $\pm 0.3$ & \textbf{80.0} $\pm \textbf{0.2}$\\
    & Pair & 78.0 $\pm 0.4$ & 78.1 $\pm 0.6$ & 78.9 $\pm 0.8$ & 79.6 $\pm 0.2$ & 78.5 $\pm 0.1$ & 79.4 $\pm 0.4$ & 77.9 $\pm 0.3$ & \textbf{80.0} $\pm \textbf{0.3}$\\
    \hline
    \multirow{2}{*}{DBLP} & Uniform & 71.0 $\pm 1.5$ & 72.4 $\pm 0.7$ & 74.9 $\pm 0.7$ & 73.1 $\pm 0.3$ & 73.5 $\pm 1.3$ & 72.8 $\pm 1.2$ & 74.2 $\pm 0.5$ & \textbf{80.8} $\pm \textbf{0.4}$\\
    & Pair & 72.5 $\pm 1.2$ & 73.4 $\pm 2.1$ & 76.3 $\pm 1.6$ & 74.4 $\pm 0.5$ & 72.7 $\pm 1.2$ & 75.4 $\pm 0.9$ & 73.6 $\pm 1.0$ & \textbf{81.1} $\pm \textbf{0.3}$\\
    \hline
    \end{tabularx}
    
    \label{tab:results}
    \vskip -1em
\end{table*}

\subsubsection{Baselines}
We compare NRGNN with representative and state-of-the-art GNNs and methods of learning with noisy labels:
\begin{itemize}[leftmargin=*]
    \item \textbf{GCN}~\cite{kipf2016semi}: GCN is a popular graph convolutional network based on spectral theory.
    \item \textbf{GIN}~\cite{xu2018powerful}: Compared with GCN, GIN could learn more powerful representations of graph structures by using multi-layer perception to process the information aggregated from the neighbors.
    \item \textbf{Self-Training}~\cite{li2018deeper}: It first trains a GCN then picks the most confident pseudo labels of GCN and puts it into the labeled node set to improve the performance of GCN.
    \item \textbf{Forward}~\cite{patrini2017making}: This is a loss correction method. It revises predictions to obtain unbiased loss on noisy training samples.
    \item \textbf{Coteaching+}~\cite{yu2019does}: This method maintains two networks to select clean samples for each other. More specifically, the small-loss samples that obtain different predictions are selected for training.

    \item \textbf{D-GNN}~\cite{nt2019learning}: It obtains a robust GNNs with backward loss correction~\cite{patrini2017making} which estimates the unbiased loss on clean labels.
    \item \textbf{CP}~\cite{zhangadversarial}: Community labels obtained by clustering node embeddings are added to train GCN. It encourages the GCN capture community information to avoid the overfitting to noisy labels.
\end{itemize}
We use GCN in Self-Training, D-GNN, CP and NRGNN to give predictions. Forward, and Coteaching+ are proposed for i.i.d data. To make a fair comparison, GCN is also adopted as backbone in these methods. 

\subsection{Node Classification with Noisy Labels}
To answer \textbf{RQ1}, we compare the proposed framework with baselines on graphs containing two types of label noise. In addition, we conduct node classification on graphs corrupted by different levels of label noise to demonstrate the effectiveness of our method.
\subsubsection{Comparisons with Baselines} Two types of label noise, i.e., uniform and pair noise, are considered for all datasets. The noise rate, i.e., the probability that a provided label is not correct, is set as 20\% for both types of label noise.  The size of the noisy labels is the same as the description in Sec~\ref{Sec:dataset}. The average results and standard deviations of 5 runs are reported in Table~\ref{tab:results}. From this table, we have the following observations:
\begin{itemize}[leftmargin=*]
    \item Both GCN and GIN perform poorly on graph with noisy and limited labels; while methods utilizing pseudo labels such as Self-Training have significantly better performance. This implies pseudo labels are helpful to alleviate the issue of learning with noisy and limited labels.
    \item Compared with Self-Training and CP which also utilize pseudo labels, the proposed NRGNN achieve higher performance under various scenarios, which is because NRGNN adopts edge predictor to add missing links between unlabeled nodes and nodes with noisy labels or pseudo labels to reduce the negative effects of the label noise. Meanwhile, these added links also help to obtain pseudo labels in higher quality . 
    \item The loss correction or sample selection based methods such as Coteaching+ and D-GNN bring limited improvements, which is due to the small training set in semi-supervised learning setting. By contrast, the proposed NRGNN outperforms these baselines by a large margin, which is because NRGNN adopts a pseudo label miner to extend the size of labeled nodes and mitigates the effects of label noise by linking unlabeled nodes and extended labeled nodes.
\end{itemize}
\begin{figure}[t]
\centering
\begin{subfigure}{0.49\columnwidth}
    \centering
    \includegraphics[width=0.92\linewidth]{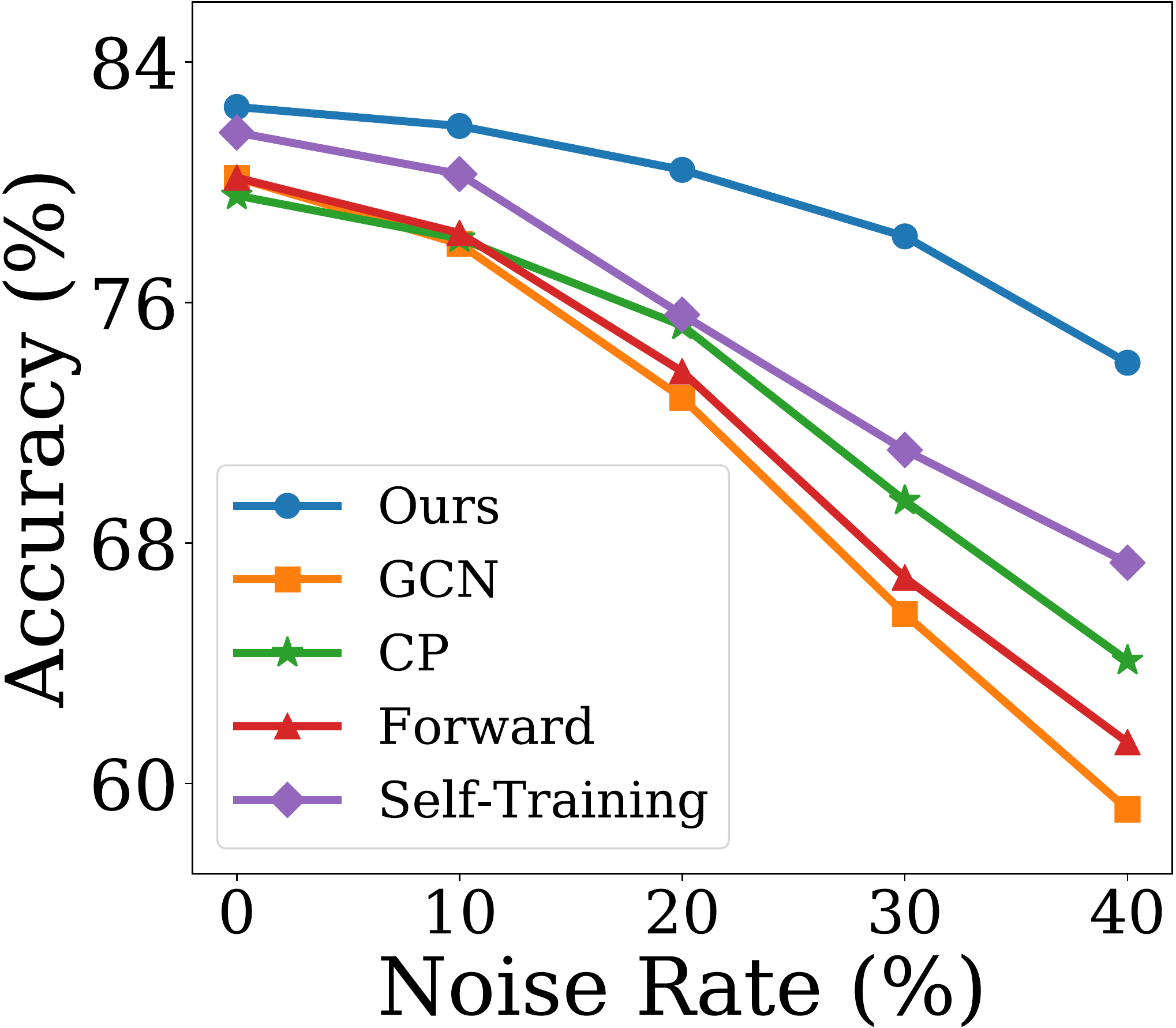} 
    \vskip -0.5em
    \caption{Uniform Noise}
\end{subfigure}
\begin{subfigure}{0.49\columnwidth}
    \centering
    \includegraphics[width=0.92\linewidth]{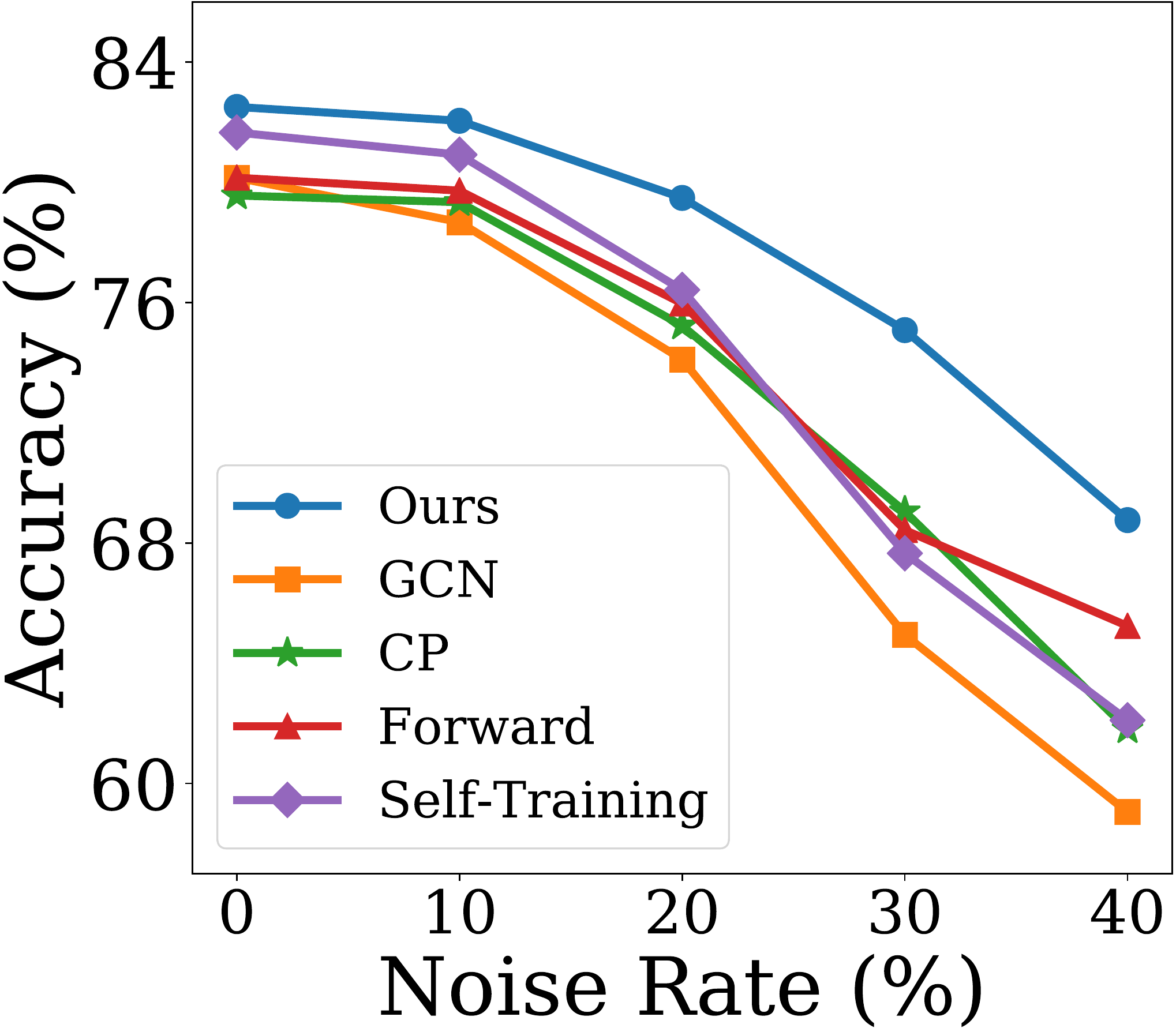} 
    \vskip -0.5em
    \caption{Pair Noise}
\end{subfigure}
\vspace{-1.3em}
\caption{Accuracy on Cora with various levels of label noise.}
\label{fig:noise_level}
\vskip -1em
\end{figure}

\subsubsection{Performance under Different Levels of Label Noise} 
To demonstrate the effectiveness of the proposed NRGNN under different levels of label noise, we vary the noise rate as $\{0\%, 10\%, \dots, 40\%\}$.  The most effective baselines in Table~\ref{tab:results} are implemented for comparisons. We only report the results on Cora, because we have similar observations for other datasets. As mentioned in Sec~\ref{Sec:dataset}, 5\% nodes are randomly sampled to compose the training set. The average performance of 5 runs is shown in Figure~\ref{fig:noise_level}. From the figure, we have the following observations:
\begin{itemize}[leftmargin=*]
    \item As the label noise level increases, the performance of all baselines drop dramatically. Though the performance of NRGNN also drops, it is more resistant to the label noise. The performance gap between NRGNN and the baselines increases when more noise exists in the labels. This implies the effectiveness of handling noisy and limited labels by extending the label set with accurate pseudo labels and adding missing links between the unlabeled and extended labeled node set.
    \item When there is little or no label noise, our proposed method still outperforms GCN and methods utilizing pseudo labels such as self-training. This is because adding high-quality edges between unlabeled nodes and extended labeled nodes could facilitate the message passing of GNNs.
\end{itemize}
\begin{figure}[t]
\centering
\begin{subfigure}{0.49\columnwidth}
    \centering
    \includegraphics[width=0.92\linewidth]{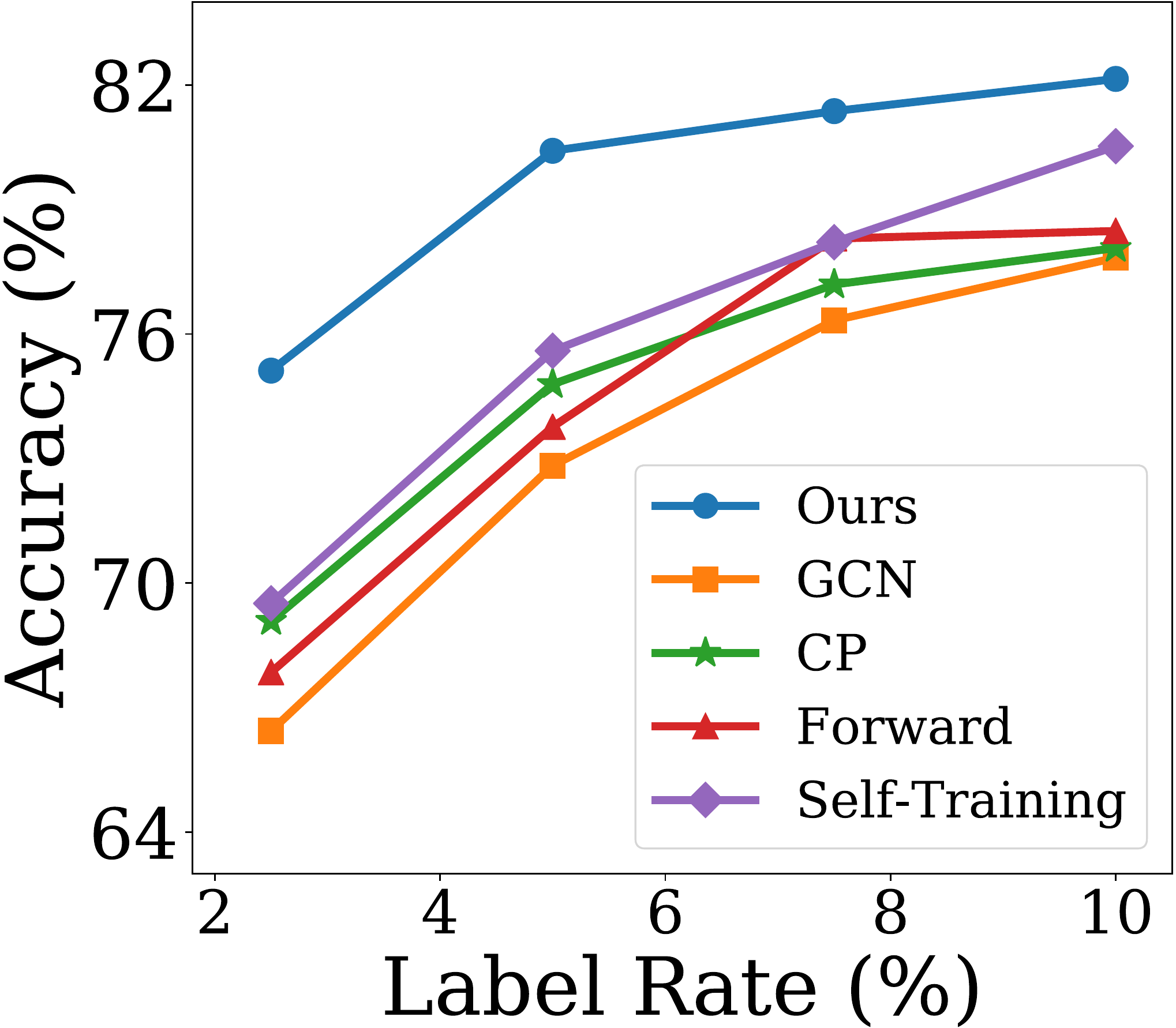} 
    \vskip -0.5em
    \caption{Uniform Noise}
\end{subfigure}
\begin{subfigure}{0.49\columnwidth}
    \centering
    \includegraphics[width=0.92\linewidth]{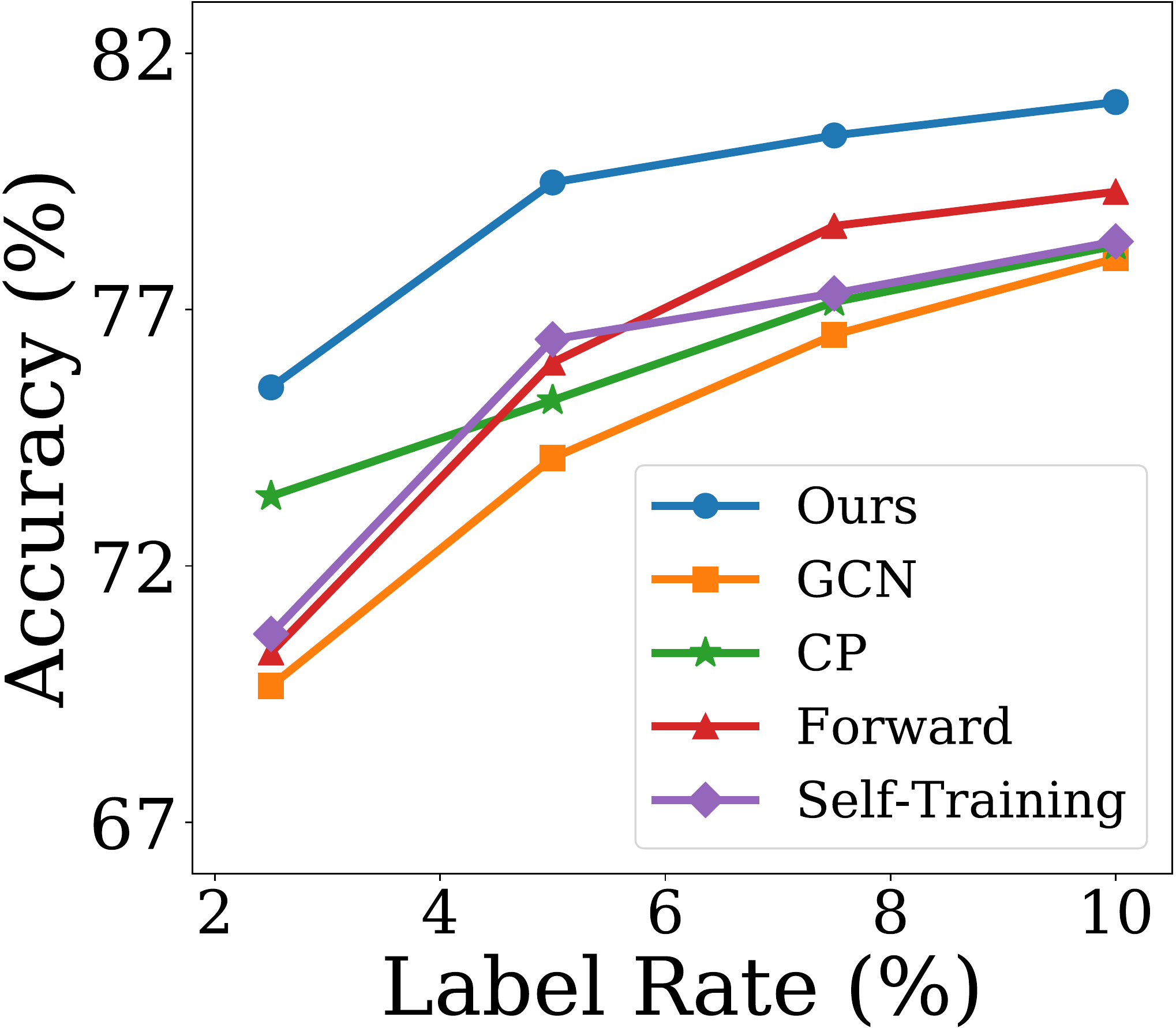} 
    \vskip -0.5em
    \caption{Pair Noise}
\end{subfigure}
\vspace{-1.3em}
\caption{Accuracy on Cora with various noisy label sizes.}
\label{fig:label_rate}
\vskip -1em
\end{figure}
\subsection{Impacts of Noisy Label Size}
In this subsection, we investigate how the size of noisy labels would affect NRGNN to answer \textbf{RQ2}. We vary the label rate, i.e., the training size, as $\{2.5\%, 5\%, 7.5\%, 10\%\}$. The noise rates of both uniform and pair noise are set as 0.2. We only report the results on  Cora in Figure~\ref{fig:label_rate} as we have similar observations on other datasets. Each experiment is run 5 times. From Figure~\ref{fig:label_rate}, we observe:
\begin{itemize}[leftmargin=*]
    \item Our proposed method brings the most significant performance improvements when the label rate is as small as 2.5\%. It indicates the effectiveness of mining accurate pseudo labels to have more supervision and benefit more from adding missing links between the unlabeled nodes and nodes with accurate pseudo labels. 
    \item With the increase of label size, the gap between our method and the baselines only decrease slightly but is still large. This is because accurate pseudo labels play a less important role when the provided noisy labels are sufficient. Though the noisy labels are sufficient, corrupted labels still degrade the performance of GNNs. The proposed NRGNN leverages the edge predictor to link more unlabeled nodes and labeled nodes to alleviate the negative effects of label noise. Thus, it can still outperform the baselines when the size of labeled nodes are large.
\end{itemize}

\subsection{Impacts of the Graph Sparsity}
\begin{figure}[t]
\centering
\begin{subfigure}{0.49\columnwidth}
    \centering
    \includegraphics[width=0.90\linewidth]{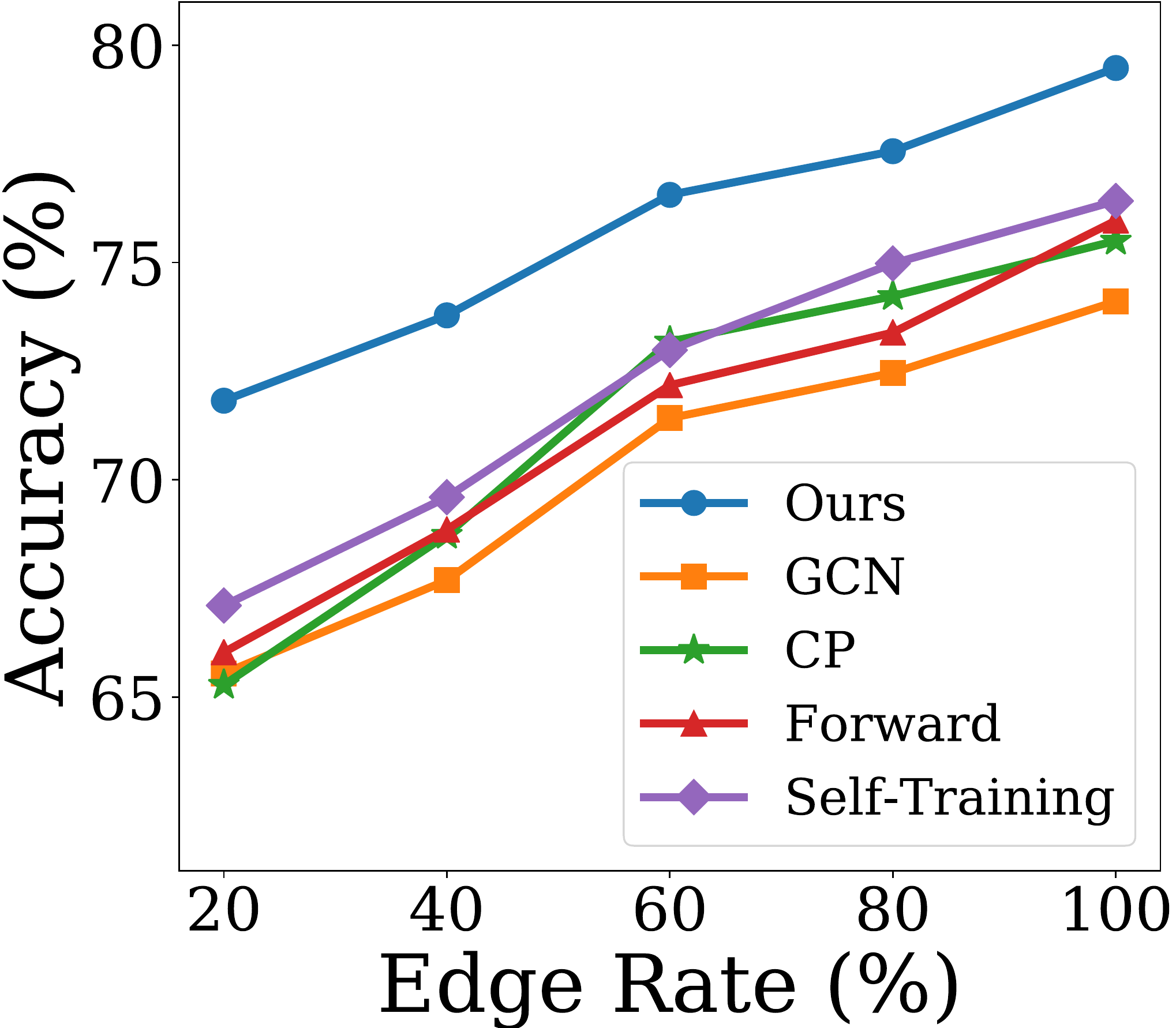} 
    \vskip -0.5em
    \caption{Cora}
\end{subfigure}
\begin{subfigure}{0.49\columnwidth}
    \centering
    \includegraphics[width=0.90\linewidth]{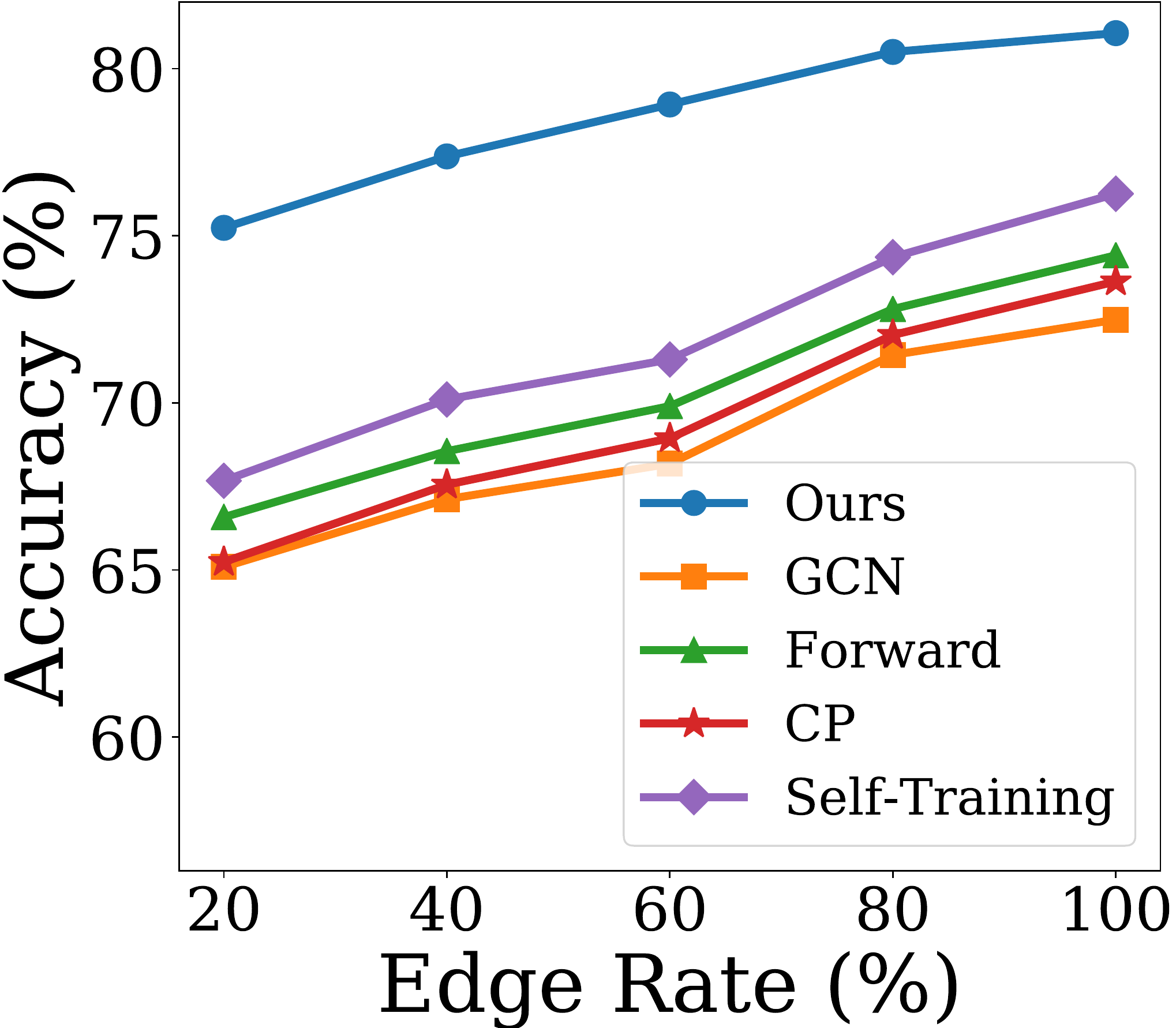} 
    \vskip -0.5em
    \caption{DBLP}
\end{subfigure}
\vspace{-1em}
\caption{Performance on graphs with different densities.}
\label{fig:density}
\vskip -1.5em
\end{figure}

The proposed NRGNN relies on an edge predictor to predict the missing links between the unlabeled nodes and labeled nodes to alleviate the effects of noisy labels. And the supervision from the adjacency matrix is utilized to have a good edge predictor. A natural question is whether NRGNN is effective when the graph is very sparse. Thus, to demonstrate that the edge predictor could learn to add useful links for robust node classification with very sparse graph, we train our model on sparse graphs obtained by randomly selecting a subset of edges in original graphs. More specifically, we vary the edge rate. i.e., the ratio of the selected edges, from 20\% to 100\% with a step of 20\%. We only report the results on Cora and DBLP corrupted by pair noise. The noise rate is set as 20\%. Average results of 5 runs are shown in Fig.~\ref{fig:density}. From the figure, we can observe that our proposed model consistently outperforms the baselines by a large margin on graphs of different sparse levels. This indicates that even with a very sparse graph, the learned edge predictor still could predict useful links between unlabeled nodes and nodes with noisy labels or pseudo labels to benefit the accurate pseudo label mining and alleviate the effects of label noise.

\subsection{Ablation Study}
\label{Sec:abl}
\begin{figure}[t]
\centering
\begin{subfigure}{0.49\columnwidth}
    \centering
    \includegraphics[width=0.90\linewidth]{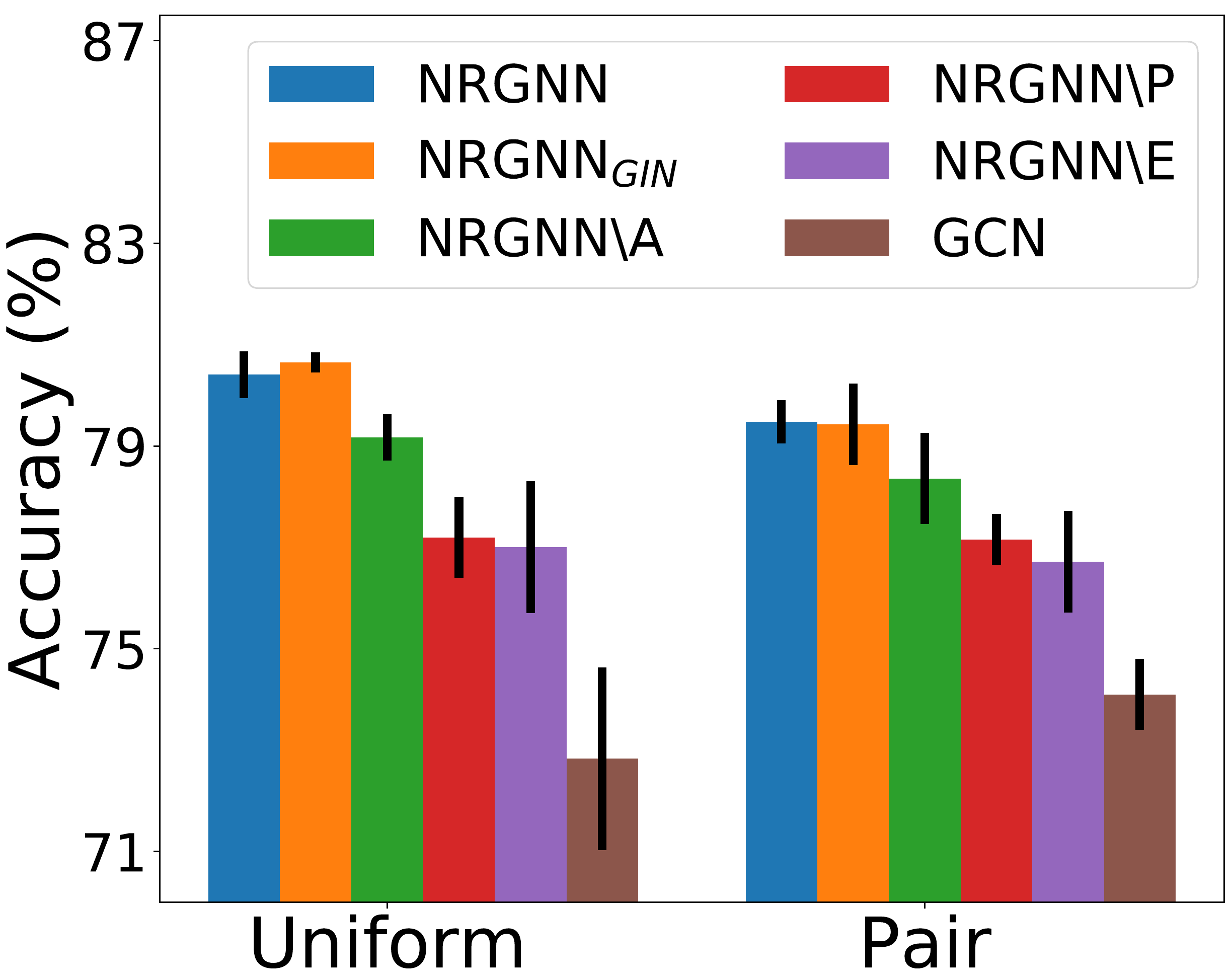} 
    \vskip -0.7em
    \caption{Cora}
\end{subfigure}
\begin{subfigure}{0.49\columnwidth}
    \centering
    \includegraphics[width=0.90\linewidth]{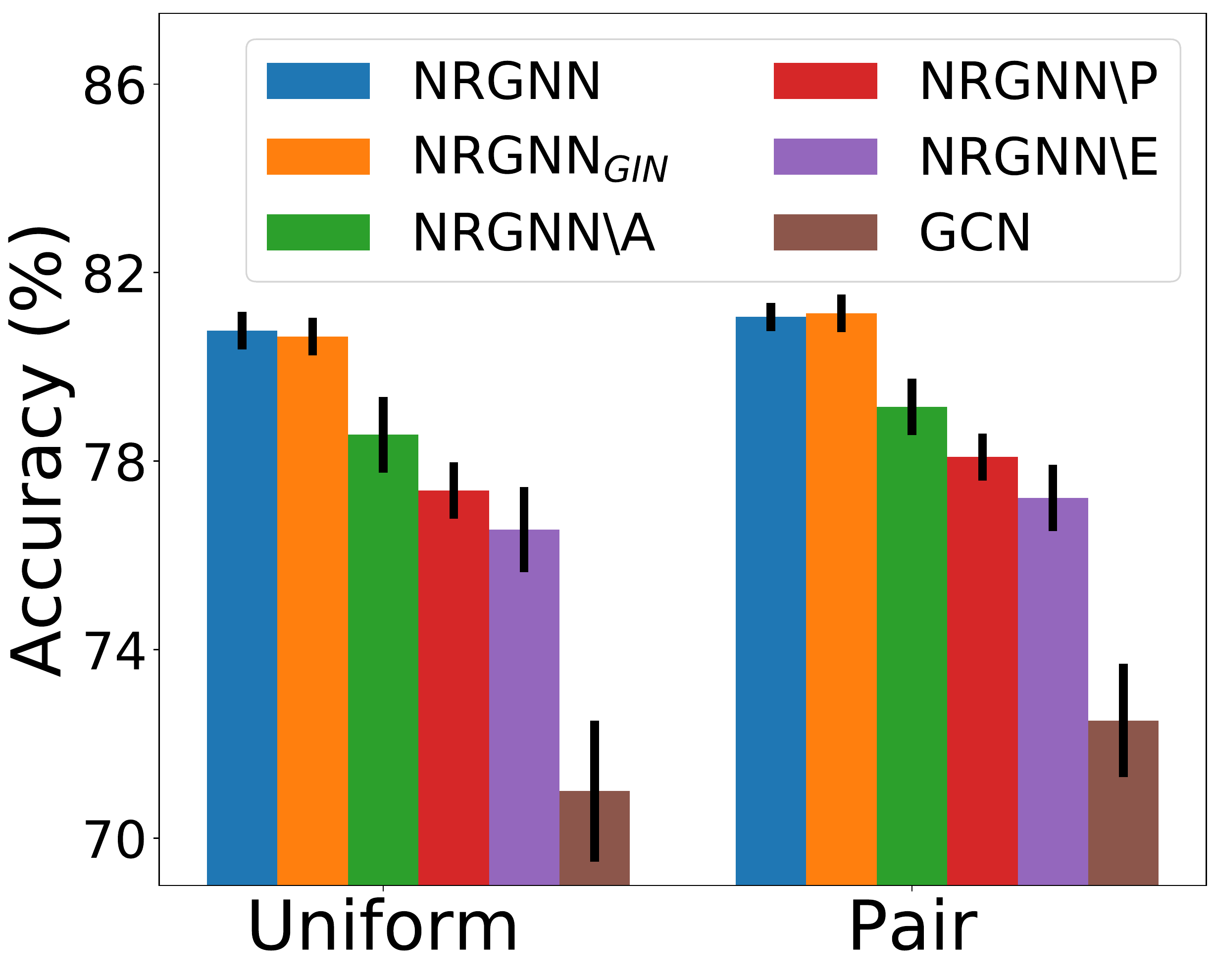} 
    \vskip -0.7em
    \caption{DBLP}
\end{subfigure}
\vspace{-1.2em}
\caption{Comparisons between NRGNN and its variants.}
\label{fig:abl}
\vskip -0.8em
\end{figure}

To answer \textbf{RQ3}, we conduct ablation study to investigate the flexibility of our proposed NRGNN and the contributions of the edge predictor and the pseudo label miner. To investigate whether various GNNs could be benefited from NRGNN, we replace the GCN classifier with a GIN classifier. More specifically, the GIN classifier is trained on the graph densified by linking similar unlabeled nodes and extended labeled nodes with the accurate pseudo labels produced by NRGNN. This variant is named as NRGNN$_{GIN}$. To demonstrate the effectiveness of the GNN-based edge predictor, we train a variant NRGNN$\backslash$E by replacing the edge predictor with cosine similarity scores of raw features.  To show the importance of the pseudo label miner, we analyze it from two aspects. Firstly, to show the contributions of pseudo labels, we train a variant NRGNN$\backslash$P which does not utilize pseudo labels. Secondly, to investigate how the quality of pseudo labels will influence the final results, we replace the accurate pseudo label miner with a GCN trained on the initial graph to obtain a variant named as NRGNN$\backslash$A. All the hyperparameters of these variants are tuned following the process described in Sec~\ref{Sec:implement}. Since we have similar observations in other datasets, we only report the performance on Cora and DBLP. The label rate is set the same as the description in Sec~\ref{Sec:dataset}. The noise rate is set as 20\%. The results of 5 runs are reported in Figure~\ref{fig:abl}. From this figure, we can observe:
\begin{itemize}[leftmargin=*]
    \item NRGNN$_{GIN}$ achieves comparable results with NRGNN, which indicates that NRGNN is flexible to various GNN backbones.
    \item The performance of NRGNN$\backslash$E is significantly worse than that of NRGNN, which shows the necessity of learning a high quality edge predictor to predict the missing links between unlabeled nodes and extended labeled nodes.
    \item The performance of NRGNN is better than that of NRGNN$\backslash$A and NRGNN$\backslash$P, which implies that pseudo label miner is helpful for learning a robust GNN with noisy and limited labels and high quality pseudo labels can bring more benefits.
    \item NRGNN$\backslash$P outperforms GCN by a large margin, which demonstrates that linking unlabeled nodes with labeled nodes can alleviate the effects of label noise.
\end{itemize}
\subsection{Hyperparameter Sensitivity Analysis}
\label{Sec:para}
\begin{figure}[t]
\centering
\begin{subfigure}{0.49\columnwidth}
    \centering
    \includegraphics[width=0.91\linewidth]{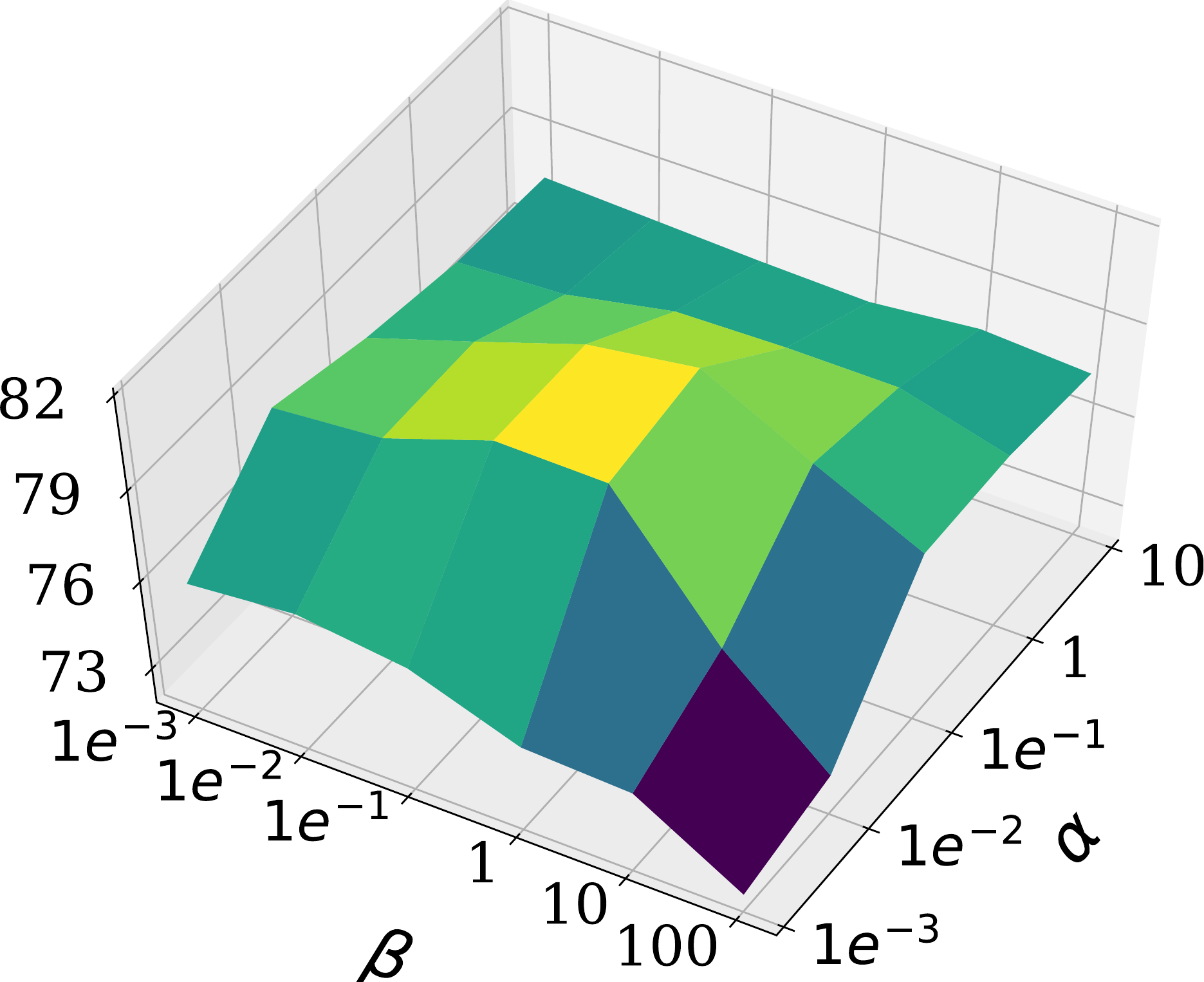} 
    \vskip -0.5em
    \caption{Uniform}
\end{subfigure}
\begin{subfigure}{0.49\columnwidth}
    \centering
    \includegraphics[width=0.91\linewidth]{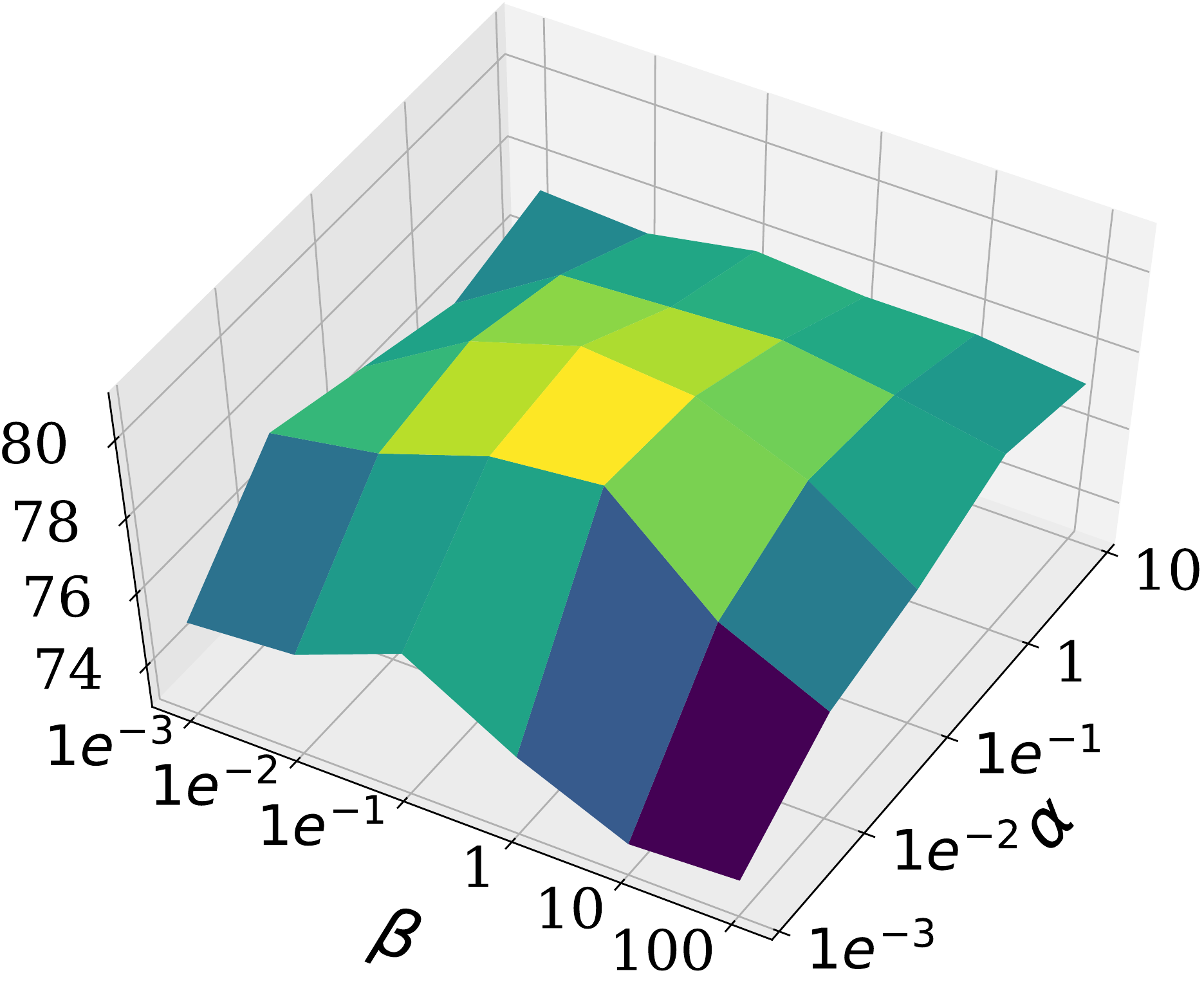} 
    \vskip -0.5em
    \caption{Pair}
\end{subfigure}
\vspace{-1.3em}
\caption{Parameter sensitivity analysis on Cora.}
\label{fig:para}
\vskip -1.5em
\end{figure}

In this subsection, we investigate how the hyperparameters $\alpha$ and $\beta$ affect the performance of NRGNN. $\alpha$ controls how well the edge predictor reconstructs the initial graph, and $\beta$ controls the learning of the pseudo label miner and its impact to the edge predictor. To explore the parameter sensitivity, we alter $\alpha$ and $\beta$ as $\{0.001, 0.01, 0.1, 1, 10\}$ and $\{0.001,0.01,0.1,1,10,100\}$, respectively. We report the results on the Cora graph corrupted by uniform and pair noise with noise rate set as 20\%. The experiments are conducted 5 times and the average results are shown in Figure~\ref{fig:para}. From the figure, we observe (i) Generally, with the increasing of $\alpha$, the performance tends to first increase and then decrease. A too small $\alpha$ would lead to a weak edge predictor while a large $\alpha$ may dominate the whole loss of NRGNN. The performance is relatively good and stable when $\alpha$ is between 0.01 and 0.1 , which eases the parameter selection for NRGNN. (ii) Similarly, with the increment of $\beta$, the performance tends to first increase and then decrease. When $\beta$ is between 0.1 and 10, the performance is relatively good. 
\section{Conclusion}
In this paper, we investigate a novel problem of semi-supervised node classification of GNN on sparsely and noisily labeled graphs. We theoretically and empirically verify the effectiveness of linking unlabeled nodes with noisily labeled nodes under mild conditions. We also show that pseudo labels could help to alleviate the limited label issue. Based on the analysis, we propose a novel framework NRGNN which utilizes an edge predictor to predict missing links for connecting unlabeled nodes with labeled nodes, and a pseudo label miner to expand the label set. With the new graph and the extended label set, a more robust GNN is trained for node classification. Experimental results on real-world datasets show the effectiveness of the proposed NRGNN on graphs with various types and levels of noise and different label and graph sparsity. Further experiments are conducted to understand the parameter sensitivity. There are several interesting directions need further investigation. First, in this paper, we mainly evaluate NRGNN under two types of noises. In practice, an adversary might on purposely attack the graph by flipping some labels to reduce the performance of GNN. We will investigate the robustness of NRGNN under adversarial label-flipping. Second, for some applications, the edges and node attributes of the given graph can also be noisy, which might affect the edge prediction. Thus, we will study how to extend NRGNN on noisy graphs with noisy labels.

\section{Acknowledgements}
This material is based upon work supported by, or in part by, the National Science Foundation (NSF) under grant  \#IIS-1909702, \#IIS1955851, and Army Research Office (ARO) under grant \#W911NF-21-1-0198. The findings and conclusions in this paper do not necessarily reflect the view of the funding agency. 

\bibliographystyle{ACM-Reference-Format}
\bibliography{acmart}

\appendix
\newpage
\renewcommand{\algorithmicrequire}{\textbf{Input:}}
\renewcommand{\algorithmicensure}{\textbf{Output:}}
\label{sec:app_alg}

\begin{algorithm}[t] 
\caption{ Training Algorithm of NRGNN.} 
\label{alg:Framwork} 
\begin{algorithmic}[1]
\REQUIRE
$\mathcal{G}=(\mathcal{V},\mathcal{E}, \mathbf{X})$, $\mathcal{Y}$, $K$, $t$, $T_p$, $\alpha$ and $\beta$.
\ENSURE $f_{\mathcal{G}}$, $f_P$ and $f_E$
\STATE Pretrain $f_P$ and $f_E$ with Eq.(\ref{eq:edge}) and Eq.(\ref{eq:loss_pseudo})
\REPEAT 
\STATE Obtain the graph $\mathbf{S}^L$ with $f_E$ by Eq.(\ref{eq:generate_graph}).
\STATE Feed $\mathbf{S}^L$ to $f_P$ to obtain pseudo labels $\mathcal{Y}_P$ by Eq.(\ref{eq:mine_pseudo})
\STATE Generate the graph $\mathbf{S}^A$ for $f_{\mathcal{G}}$ with $f_E$ by Eq.(\ref{eq:generate_graph_all})
\STATE Jointly optimize the parameters of $f_{\mathcal{G}}$, $f_P$ and $f_E$ by Eq.(\ref{eq:loss_final})
\UNTIL convergence
\RETURN $f_{\mathcal{G}}$, $f_P$ and $f_E$
\end{algorithmic}
\end{algorithm}
\section{Training Algorithm}
\label{app:alg}
The training algorithm of NRGNN is shown in Algorithm~\ref{alg:Framwork}. In line 1, edge predictor $f_E$ and accurate pseudo label miner $f_P$ will be pretrained with Eq.(\ref{eq:edge}) and Eq.(\ref{eq:loss_pseudo}). In line 2, we generate $\mathbf{S}^P$ for $f_P$ with $f_E$. Then, the accurate pseudo labels could be obtained. In line 5, the graph $\mathbf{S}^A$ which linking nodes with similar extended labeled nodes is obtained for $f_\mathcal{G}$ to make robust predictions. Finally, $f_\mathcal{G}$, $f_E$ and $f_P$ will be jointly trained with an Adam optimizer~\cite{kingma2014adam} with the learning rate set as 0.001.
\section{Proof of Theorem~\ref{theorem:labeled}}
\label{app:proof_labeled}
\begin{proof} 
The predicted probability that node $v_u$ belongs to the class $c$ could be rewritten to the following format:
\begin{equation}
    y_{uc} = \frac{1}{m+n}(\sum_{v_a\in \mathcal{V}_a}s_{ac} + \sum_{v_l\in \mathcal{V}_n}s_{lc}),
\end{equation}
where $\mathcal{V}_a$ denotes the unlabeled neighbors of $v_u$, $\mathcal{V}_n$ denotes the linked nodes with noisy labels.
Let $p_t$ denotes the probability that a node belonging to class $c$ is assigned to label $c$, and $p_f$ denotes the probability that a node not belonging to class $c$ is assigned to label $c$. Then average value of $y_{uc}$ would be:
\begin{equation}
\begin{aligned}
     \mathbb{E}{(y_{uc})} = & \frac{n}{m+n}\mathbb{E}(s_{ac}) + \frac{(hp_{t} + (1-h)p_{f}))m}{m+n}\mathbb{E}(s_{bc}) \\
                            & + \frac{(h(1-p_t)+(1-h)(1-p_f))m}{m+n}\mathbb{E}(s_{dc}),
\end{aligned}
\label{eq:E(y)}
\end{equation}
where $s_{ac}$ corresponds to the unlabeled node $v_a \in \mathcal{V}_U$, $s_{bc}$ corresponds to the labeled node $v_b \in \mathcal{V}_L$ whose provided label is $c$, and $s_{dc}$ corresponds to the labeled node $v_d \in \mathcal{V}_L$ whose provided label is not $c$. Since $p_t > p_f$, we could have $p=(hp_t+(1-h)p_f) < p_t$. And Eq.(\ref{eq:E(y)}) could be rewritten to:
\begin{equation}
    \mathbb{E}(y_{uc}) = \frac{n\mathbb{E}(s_{ac}) + pm\mathbb{E}(s_{bc}) + (1-p)m \mathbb{E}(s_{dc})}{m+n}.
\end{equation}
If we further link $v_u$ with $k$ labeled nodes which belong to $c$. Then we could obtain the corresponding predicted probability $y_{uc}^k$. The expectation of $y_{uc}^k$ can be written as:
\begin{equation}
\begin{aligned}
     \mathbb{E}(y_{uc}^k) = \frac{m+n}{m+n+k}\mathbb{E}(y_{uc}) + \frac{kp_t \mathbb{E}(s_{bc}) + k(1-p_t)\mathbb{E}(s_{dc})}{m+n+k}.
\end{aligned}
\end{equation}
Since $p<p_t$ and $\mathbb{E}(s_{bc}) > \mathbb{E}(s_{ac})>\mathbb{E}(s_{dc})$, we can derive that
\begin{equation}
    p_t\mathbb{E}(s_{bc})+(1-p_t)\mathbb{E}(s_{dc}) >
    p\mathbb{E}(s_{bc}) + (1-p)\mathbb{E}(s_{dc}).
    \label{eq:a}
\end{equation}
When $p_t > \frac{\mathbb{E}(s_{ac})-\mathbb{E}(s_{dc})}{\mathbb{E}(s_{bc})-\mathbb{E}(s_{dc})}$, we could have
\begin{equation}
    p_t\mathbb{E}(s_{bc})+(1-p_t)\mathbb{E}(s_{dc})>\mathbb{E}(s_{ac}).
    \label{eq:b}
\end{equation}
Combining Eq.(\ref{eq:a}) and Eq.(\ref{eq:b}), we can derive
\begin{equation}
    p_t\mathbb{E}(s_{bc})+(1-p_t)\mathbb{E}(s_{dc})>\mathbb{E}(y_{uc}).
\end{equation}
Therefore, we could conclude  $\mathbb{E}(y_{uc}^k)>\mathbb{E}(y_{uc})$. And with the increasing of $k$, the predicted probability that node $v_u$ belonging to class $c$ would increase.
\end{proof}

\section{Proof of Theorem~\ref{theorem:pseudo}}
\label{app:proof_pseudo}
\begin{proof}
The average value of $y_{uc}$ could be written as:
\begin{equation}
     \mathbb{E}(y_{uc}) = \frac{n\mathbb{E}(s_{ac}) + pm\mathbb{E}(s_{bc}) + (1-p)m \mathbb{E}(s_{dc})}{m+n},
\end{equation}
Since $\mathbb{E}(s_{pc}) > \mathbb{E}(s_{ac})$ and $\mathbb{E}(s_{pc}) > p\mathbb{E}(s_{bc})+(1-p)\mathbb{E}(s_{dc}))$, then we could have $\mathbb{E}(s_{pc}) > \mathbb{E}(y_{uc})$. Therefore, the expectation of $y_{uc}$ after linking $k$ nodes with pseudo labels would be:
\begin{equation}
    \mathbb{E}(y_{uc}^k) = \frac{m+n}{m+n+k}\mathbb{E}(y_{uc}) + \frac{k}{m+n+k}\mathbb{E}(s_{pc}).
\end{equation}
Since $\mathbb{E}(s_{pc}) > \mathbb{E}(y_{uc})$, we could conclude that with the increasing of $k$, $\mathbb{E}(y_{uc}^k)$ would be higher.
\end{proof}
\begin{table}[t]
    \small
    \caption{Statistics of datasets.}
    \vskip-1.5em
    \centering
    \begin{tabularx}{0.95\linewidth}{p{0.27\linewidth}XXXX}
    \toprule
         & Cora & Citeseer & Pubmed  & DBLP \\
    \midrule
    \# of nodes & 2,485 & 2,110 & 19,717 & 17,716\\
    \# of edges & 5,068 & 3,668 & 44,338 & 52,867\\
    \# of features & 1,433 & 3,703 & 500 & 1,639\\
    \# of classes & 7 & 6 & 3 & 4\\
    \bottomrule
    \end{tabularx}
    \label{tab:dataset}
\end{table}
\end{document}